%% file: paper.tex
\documentclass{article}
\usepackage{iclr2017,times}
\usepackage{hyperref}
\usepackage{url}

\RequirePackage{snapshot}
\usepackage[utf8]{inputenc}
\usepackage[T1]{fontenc}
\usepackage{url}
\usepackage{booktabs}
\usepackage{amsfonts}
\usepackage{amsmath}
\usepackage{nicefrac}
\usepackage{microtype}
\usepackage[sort]{cite}
\usepackage{tikz}
\usepackage{subcaption}
\usepackage{comment}
\usepackage[export]{adjustbox}
\usepackage{mathtools}
\usepackage{multirow}
\usepackage{thmtools, thm-restate}
\usepackage{xspace}
\usepackage[titletoc,title]{appendix}
\usetikzlibrary{shapes, arrows, shadows, fit, backgrounds, calc}

\usepackage{xcolor}
\definecolor{Gray}{RGB}{150, 150, 150}
\definecolor{Plum}{RGB}{142, 69, 133}
\definecolor{Red3}{HTML}{a40000}
\definecolor{Green3}{HTML}{4e9a06}
\newcommand{\gray}[1]{\textcolor{Gray}{#1}}

\newcommand{\proof}{\textit{Proof}.\xspace}

\newcommand{\expect}[2]{\mathbb{E}_{#1}\left[#2\right]}

\newcommand{\ldeltafunc}[1]{\mathbf{1}_{\left[#1\right]}}
\newcommand{\logp}[1]{\log\left(#1\right)}
\newcommand{\jsd}[2]{\mathrm{D_{JS}}\left(\left.#1\,\middle|\middle|\,#2\right.\right)}
\newcommand{\kld}[2]{\mathrm{D_{KL}}\left(\left.#1\,\middle|\middle|\,#2\right.\right)}
\newcommand{\defeq}{\vcentcolon=}

\renewcommand{\d}{\mathop{}\!\mathrm{d}}

\def\smallint{\begingroup\textstyle \int\endgroup}

\renewcommand{\vec}[1]{\mathbf{#1}}
\newcommand{\x}{\vec{x}}

\newcommand{\X}{\vec{X}}
\newcommand{\OX}{\Omega_\X}
\newcommand{\hOX}{\hat{\Omega}_\X}

\newcommand{\z}{\vec{z}}

\newcommand{\Z}{\vec{Z}}

\newcommand{\OZ}{\Omega_\Z}
\newcommand{\hOZ}{\hat{\Omega}_\Z}

\DeclareMathOperator*{\argmax}{arg\,max}
\DeclareMathOperator*{\supp}{supp}

\newcommand{\reffig}[1]{Figure~\ref{fig:#1}}

\newcommand{\refsec}[1]{Section~\ref{sec:#1}}

\newcommand{\reftbl}[1]{Table~\ref{tbl:#1}}

\newcommand{\lblfig}[1]{\label{fig:#1}}
\newcommand{\lblsec}[1]{\label{sec:#1}}

\newcommand{\lbltbl}[1]{\label{tbl:#1}}

\title{Adversarial Feature Learning}

\author{Jeff Donahue\\
\texttt{jdonahue@cs.berkeley.edu}\\
Computer Science Division\\
University of California, Berkeley
\And
Philipp Kr\"ahenb\"uhl\\
\texttt{philkr@utexas.edu}\\
Department of Computer Science\\
University of Texas, Austin
\And
Trevor Darrell\\
\texttt{trevor@eecs.berkeley.edu}\\
Computer Science Division\\
University of California, Berkeley}

\iclrfinalcopy

\begin{document}

\maketitle

\begin{abstract}
  The ability of the Generative Adversarial Networks (GANs) framework
  to learn generative models mapping from simple latent distributions
  to arbitrarily complex data distributions has been demonstrated
  empirically, with compelling results showing
  that the latent space of such generators captures semantic variation
  in the data distribution.
  Intuitively, models trained to predict these semantic latent representations
  given data may serve as useful feature representations
  for auxiliary problems where semantics are relevant.
  However, in their existing form, GANs have no means of learning the inverse mapping --
  projecting data back into the latent space.
  We propose Bidirectional Generative Adversarial Networks (BiGANs) as a means of
  learning this inverse mapping,
  and demonstrate that the resulting learned feature representation
  is useful for auxiliary supervised discrimination tasks, competitive with contemporary
  approaches to unsupervised and self-supervised feature learning.
\end{abstract}

\section{Introduction}
\lblsec{intro}
Deep convolutional networks (convnets)
have become a staple of the modern computer vision pipeline.
After training these models on a massive database of image-label pairs like ImageNet~\citep{imagenet}, the network easily adapts to a variety of similar visual tasks, achieving impressive results on image classification~\citep{decaf,zf,astounding} or localization~\citep{rcnn,fcn} tasks.
In other perceptual domains such as natural language processing or speech recognition, deep networks have proven highly effective as well~\citep{bengiotranslate,orioltranslate,oriolparse,graves}.
However, all of these recent results rely on a supervisory signal from large-scale databases of hand-labeled data, ignoring much of the useful information present in the structure of the data itself.

Meanwhile, Generative Adversarial Networks (GANs)~\citep{gan} have emerged as a powerful framework for learning generative models of arbitrarily complex data distributions.
The GAN framework learns a \textit{generator} mapping samples from an arbitrary latent distribution to data, as well as an adversarial \textit{discriminator} which tries to distinguish between real and generated samples as accurately as possible.
The generator's goal is to ``fool'' the discriminator by producing samples which are as close to real data as possible.
When trained on databases of natural images, GANs produce impressive results~\citep{dcgan,lapgan}.

Interpolations in the latent space of the generator produce smooth and plausible semantic variations,
and certain directions in this space correspond to particular semantic attributes along which the data distribution varies.
For example, \citet{dcgan} showed that a GAN trained on a database of human faces learns to associate particular latent directions with gender and the presence of eyeglasses.

A natural question arises from this ostensible ``semantic juice'' flowing through the weights of generators learned using the GAN framework:
can GANs be used for unsupervised learning of rich feature representations for arbitrary data distributions?
An obvious issue with doing so is that the generator maps latent samples to generated data,
but the framework does not include an \textit{inverse} mapping from data to latent representation.

Hence, we propose a novel unsupervised feature learning framework,
\textit{Bidirectional Generative Adversarial Networks} (BiGAN).
The overall model is depicted in \reffig{concept}.
In short, in addition to the generator $G$ from the standard GAN framework~\citep{gan},
BiGAN includes an \textit{encoder} $E$ which maps data $\x$ to latent representations $\z$.
The BiGAN discriminator $D$ discriminates not only in data space ($\x$ versus $G(\z)$),
but jointly in data and latent space (tuples $(\x, E(\x))$ versus $(G(\z), \z)$),
where the latent component is either an encoder output $E(\x)$ or a generator input $\z$.

It may not be obvious from this description that the BiGAN encoder $E$ should learn to invert the generator $G$.
The two modules cannot directly ``communicate'' with one another:
the encoder never ``sees'' generator outputs ($E(G(\z))$ is not computed), and vice versa.
Yet, in Section~\ref{sec:bigan}, we will both argue intuitively and formally prove that the encoder and generator
must learn to invert one another in order to fool the BiGAN discriminator.

Because the BiGAN encoder learns to predict features $\z$ given data $\x$,
and prior work on GANs has demonstrated that these features capture semantic attributes of the data,
we hypothesize that a trained BiGAN encoder may serve as a useful feature representation for related semantic tasks,
in the same way that fully supervised visual models trained to predict semantic ``labels'' given images
serve as powerful feature representations for related visual tasks.
In this context, a latent representation $\z$ may be thought of as a ``label'' for $\x$,
but one which came for ``free,'' without the need for supervision.

An alternative approach to learning the inverse mapping from data to latent representation
is to directly model $p(\z|G(\z))$,
predicting generator input $\z$ given generated data $G(\z)$.
We'll refer to this alternative as a \textit{latent regressor},
later arguing (Section~\ref{sec:baselines})
that the BiGAN encoder may be preferable in a feature learning context,
as well as comparing the approaches empirically.

BiGANs are a robust and highly generic approach to unsupervised feature learning,
making no assumptions about the structure or type of data to which they are applied,
as our theoretical results will demonstrate.
Our empirical studies will show that despite their generality,
BiGANs are competitive with contemporary approaches to self-supervised and weakly supervised feature learning
designed specifically for a notoriously complex data distribution -- natural images.

\citet{ali} independently proposed an identical model in their concurrent work,
exploring the case of a stochastic encoder $E$ and the ability of such models to learn in a semi-supervised setting.

\input{concept}

\section{Preliminaries}
Let $p_\X(\x)$ be the distribution of our data for $\x \in \OX$ (e.g. natural images).
The goal of generative modeling is capture this data distribution using a probabilistic model.
Unfortunately, exact modeling of this probability density function is computationally intractable~\citep{hinton2006fast,salakhutdinov2009deep} for all but the most trivial models.
Generative Adversarial Networks (GANs)~\citep{gan} instead model the data distribution as a transformation of a fixed latent distribution $p_\Z(\z)$ for $\z \in \OZ$.
This transformation, called a \textit{generator}, is expressed as a deterministic feed forward network $G:\OZ \to \OX$ with $p_G(\x | \z) = \delta\left(\x - G(\z)\right)$ and $p_G(\x) = \expect{\z \sim p_\Z}{p_G(\x | \z)}$.
The goal is to train a generator such that $p_G(\x) \approx p_\X(\x)$.

The GAN framework trains a generator, such that no discriminative model $D : \OX \mapsto [0, 1]$ can distinguish samples of the data distribution from samples of the generative distribution.
Both generator and discriminator are learned using the adversarial (minimax) objective $\min\limits_G \max\limits_D V(D, G)$, where
\begin{equation}
V(D, G)
\defeq
\expect{\x \sim p_\X}{
  \log D(\x)
}
+
\underbrace{
  \expect{\x \sim p_G}{
    \logp{1 - D(\x)}
  }
}_{
  \mathbb{E}_{\z \sim p_\Z}[\logp{1 - D(G(\z))}]
}
\label{eq:generalganobj}
\end{equation}

\citet{gan} showed that for an ideal discriminator the objective $C(G) \defeq \max_D V(D, G)$ is equivalent to the Jensen-Shannon divergence between the two distributions $p_G$ and $p_\X$.

The adversarial objective~\ref{eq:generalganobj} does not directly lend itself to an efficient optimization, as each step in the generator $G$ requires a full discriminator $D$ to be learned.
Furthermore, a perfect discriminator no longer provides any gradient information to the generator, as the gradient of any global or local maximum of $V(D, G)$ is $0$.
To provide a strong gradient signal nonetheless, \citet{gan} slightly alter the objective between generator and discriminator updates, while keeping the same fixed point characteristics.
They also propose to optimize (\ref{eq:generalganobj}) using an alternating optimization switching between updates to the generator and discriminator.
While this optimization is not guaranteed to converge, empirically it works well if the discriminator and generator are well balanced.

Despite the empirical strength of GANs as generative models of arbitrary data distributions,
it is not clear how they can be applied as an unsupervised feature representation.
One possibility for learning such representations is to learn an inverse mapping regressing from generated data $G(\z)$ back to the latent input $\z$.
However, unless the generator perfectly models the data distribution $p_\X$,
a nearly impossible objective for a complex data distribution such as that of high-resolution natural images,
this idea may prove insufficient.

\section{Bidirectional Generative Adversarial Networks}
\lblsec{bigan}
In Bidirectional Generative Adversarial Networks (BiGANs) we not only train a generator, but additionally train an encoder $E:\OX \to \OZ$.
The encoder induces a distribution $p_E(\z | \x) = \delta(\z - E(\x))$ mapping data points $\x$ into the latent feature space of the generative model.
The discriminator is also modified to take input from the latent space, predicting $P_D(Y | \x, \z)$,
where $Y=1$ if $\x$ is real (sampled from the real data distribution $p_\X$),
and $Y=0$ if $\x$ is generated (the output of $G(\z), \z \sim p_\Z$).

The BiGAN training objective is defined as a minimax objective
\begin{equation}
\min\limits_{G, E}
\max\limits_D
V(D, E, G)
\end{equation}
where
\begin{align}
V(D, E, G)
\defeq
\mathbb{E}_{\x \sim p_\X}\big[
  \underbrace{
    \expect{\z \sim p_E(\cdot | \x)}{
      \log D(\x, \z)
    }
  }_{
    \log D(\x, E(\x))
  }
\big]
+
\mathbb{E}_{\z \sim p_\Z}\big[
  \underbrace{
    \expect{\x \sim p_G(\cdot | \z)}{
      \logp{1 - D(\x, \z)}
    }
  }_{
    \logp{1 - D(G(\z), \z)}
  }
\big].
\label{eq:vdge}
\end{align}
We optimize this minimax objective using the same alternating gradient based optimization as \citet{gan}.
See \refsec{learning} for details.

BiGANs share many of the theoretical properties of GANs~\citep{gan}, while additionally guaranteeing that at the global optimum, $G$ and $E$ are each other's inverse.
BiGANs are also closely related to autoencoders with an $\ell_0$ loss function.
In the following sections we highlight some of the appealing theoretical properties of BiGANs.

\paragraph{Definitions}
\label{sec:defs}
Let $p_{G\Z}(\x, \z) \defeq p_G(\x | \z) p_\Z(\z)$ and $p_{E\X}(\x, \z) \defeq p_E(\z | \x) p_\X(\x)$
be the joint distributions modeled by the generator and encoder respectively.
$\Omega \defeq \OX \times \OZ$ is the joint latent and data space. For a region $R \subseteq \Omega$,
\begin{align*}
  P_{E\X}(R)
  &\defeq \smallint_{\Omega} p_{E\X}(\x, \z) \ldeltafunc{(\x, \z) \in R} \d(\x, \z)
  = \smallint_{\OX} p_\X(\x) \smallint_{\OZ} p_E(\z | \x) \ldeltafunc{(\x, \z) \in R} \d\z \d\x \\
  P_{G\Z}(R)
  &\defeq \smallint_{\Omega} p_{G\Z}(\x, \z) \ldeltafunc{(\x, \z) \in R} \d(\x, \z)
  = \smallint_{\OZ} p_\Z(\z) \smallint_{\OX} p_G(\x | \z) \ldeltafunc{(\x, \z) \in R} \d\x \d\z
\end{align*}
are probability measures over that region.
We also define
\begin{align*}
P_{\X}(R_\X)
&\defeq \smallint_{\OX} p_\X(\x) \ldeltafunc{\x \in R_\X} \d\x
&
P_{\Z}(R_\Z)
&\defeq \smallint_{\OZ} p_\Z(\z) \ldeltafunc{\z \in R_\Z} \d\z
\end{align*}
as measures over regions $R_\X \subseteq \OX$ and $R_\Z \subseteq \OZ$.
We refer to the set of features and data samples in the support of $P_\X$ and $P_\Z$ as
$ \hOX \defeq \supp(P_\X) $ and $ \hOZ \defeq \supp(P_\Z) $ respectively.
$\kld{P}{Q}$
and
$\jsd{P}{Q}$
respectively denote the
Kullback-Leibler (KL) and Jensen-Shannon divergences
between probability measures
$P$ and $Q$. By definition,
\begin{align*}
\kld{P}{Q}
&\defeq
\expect{\x \sim P}{\log f_{PQ}(\x)}
\\
\jsd{P}{Q}
&\defeq
\tfrac{1}{2}\left(
\kld{P}{\tfrac{P+Q}{2}}
+
\kld{Q}{\tfrac{P+Q}{2}}
\right),
\end{align*}
where $
f_{PQ} \defeq
\tfrac{\d P}{\d Q}
$ is the Radon-Nikodym (RN) derivative of measure $P$ with respect to measure $Q$, with the defining property that
$
P(R)
=
\int_R f_{PQ} \d Q
$.
The RN derivative $f_{PQ} : \Omega \mapsto \mathbb{R}_{\ge 0}$ is defined for any measures $P$ and $Q$ on space $\Omega$ such that $P$ is absolutely continuous with respect to $Q$: i.e., for any $R \subseteq \Omega$, $P(R)>0 \implies Q(R)>0$.

\newtheorem{theorem}{Theorem}
\newtheorem{corollary}{Corollary}
\newtheorem{proposition}{Proposition}
\newtheorem{lemma}{Lemma}

\subsection{Optimal discriminator, generator, \& encoder}
We start by characterizing the optimal discriminator for any generator and encoder,
following~\citet{gan}.
This optimal discriminator then allows us to reformulate objective \eqref{eq:vdge},
and show that it reduces to the Jensen-Shannon divergence between the joint distributions $P_{E\X}$ and $P_{G\Z}$.
\begin{restatable}{proposition}{optdiscrim}
\label{eq:optdiscrim}
For any $E$ and $G$, the optimal discriminator $
D^*_{EG} \defeq
\argmax_D V(D, E, G)
$ is the Radon-Nikodym derivative $
f_{EG} \defeq
\frac{\d P_{E\X}}{\d (P_{E\X} + P_{G\Z})}
: \Omega \mapsto [0, 1]
$
of measure $
P_{E\X}
$ with respect to measure $
P_{E\X} + P_{G\Z}
$.
\end{restatable}
\proof Given in Appendix~\ref{sec:optdiscrim}.

This optimal discriminator now allows us to characterize the optimal generator and encoder.

\begin{restatable}{proposition}{cegjsd}
\label{cegjsd}
The encoder and generator's objective for an optimal discriminator $
C(E, G)
\defeq
\max_D
V(D, E, G)
=
V(D^*_{EG}, E, G)
$ can be rewritten in terms of the Jensen-Shannon divergence between measures $P_{E\X}$ and $P_{G\Z}$ as $
C(E, G)
=
2 \, \jsd{P_{E\X}}{P_{G\Z}} - \log 4
$.
\end{restatable}
\proof Given in Appendix~\ref{sec:cegjsd}.

\begin{theorem}
The global minimum of $
C(E, G)
$
is achieved if and only if $
P_{E\X} = P_{G\Z}
$.
At that point, $C(E, G) = - \log 4$ and $D^*_{EG} = \frac{1}{2}$.
\label{thm:optgen}
\end{theorem}

\proof
From Proposition~\ref{cegjsd}, we have that $
C(E, G) =
2 \, \jsd{P_{E\X}}{P_{G\Z}} - \log 4
$.
The Jensen-Shannon divergence $\jsd{P}{Q} \ge 0$ for any $P$ and $Q$,
and $\jsd{P}{Q}=0$ if and only if $P = Q$.
Therefore, the global minimum of $
C(E, G)
$ occurs if and only if $
P_{E\X} = P_{G\Z}
$,
and at this point the value is $
C(E, G)
=
- \log 4
$.
Finally, $
P_{E\X} = P_{G\Z}
$ implies that the optimal discriminator is chance: $
D^*_{EG} =
\tfrac{\d P_{E\X}}{\d(P_{E\X} + P_{G\Z})} =
\tfrac{\d P_{E\X}}{2 \d P_{E\X}} =
\tfrac{1}{2}
$.
$\square$

The optimal discriminator, encoder, and generator of BiGAN are similar to the optimal discriminator and generator of the GAN framework~\citep{gan}.
However, an important difference is that BiGAN optimizes a Jensen-Shannon divergence between a joint distribution over both data $\X$ and latent features $\Z$.
This joint divergence allows us to further characterize properties of $G$ and $E$, as shown below.

\subsection{Optimal generator \& encoder are inverses}
We first present an intuitive argument that, in order to ``fool'' a perfect discriminator, a deterministic BiGAN encoder and generator must invert each other.
(Later we will formally state and prove this property.)
Consider a BiGAN discriminator input pair $(\x, \z)$.
Due to the sampling procedure, $(\x, \z)$ must satisfy at least one of the following two properties:
\begin{align*}
\text{(a) }
  \x \in \hOX
  \, \land \,
  E(\x) = \z
& & &
\text{(b) }
  \z \in \hOZ
  \, \land \,
  G(\z) = \x
\end{align*}
If \emph{only} one of these properties is satisfied, a perfect discriminator can infer the source of $(\x, \z)$ with certainty:
if only (a) is satisfied, $(\x, \z)$ must be an encoder pair $(\x, E(\x))$ and $D^*_{EG}(\x, \z) = 1$;
if only (b) is satisfied, $(\x, \z)$ must be a generator pair $(G(\z), \z)$ and $D^*_{EG}(\x, \z) = 0$.

Therefore, in order to fool a perfect discriminator at $(\x, \z)$ (so that $0 < D^*_{EG}(\x, \z) < 1$), $E$ and $G$ must satisfy \emph{both} (a) and (b).
In this case, we can substitute the equality $E(\x)=\z$ required by (a) into the equality $G(\z)=\x$ required by (b), and vice versa,
giving the inversion properties $\x = G(E(\x))$ and $\z = E(G(\z))$.

Formally, we show in Theorem~\ref{th:inverse} that the optimal generator and encoder invert one another almost everywhere on the support $\hOX$ and $\hOZ$ of $P_\X$ and $P_\Z$.
\begin{restatable}{theorem}{thinv}
  If $E$ and $G$ are an optimal encoder and generator, then $E = G^{-1}$ almost everywhere; that is,
  $G(E(\x))=\x$ for $P_\X$-almost every $\x \in \OX$, and
  $E(G(\z))=\z$ for $P_\Z$-almost every $\z \in \OZ$.
  \label{th:inverse}
\end{restatable}
\proof Given in Appendix~\ref{sec:geinv}.

While Theorem \ref{th:inverse} characterizes the encoder and decoder at their optimum,
due to the non-convex nature of the optimization, this optimum might never be reached.
Experimentally, \refsec{eval} shows that on standard datasets,
the two are approximate inverses; however, they are rarely exact inverses.
It is thus also interesting to show what objective BiGAN optimizes in terms of $E$ and $G$.
Next we show that BiGANs are closely related to autoencoders with an $\ell_0$ loss function.

\subsection{Relationship to autoencoders}

As argued in Section~\ref{sec:intro}, a model trained to predict features $\z$ given data $\x$ should learn useful semantic representations.
Here we show that the BiGAN objective forces the encoder $E$ to do exactly this:
in order to fool the discriminator at a particular $\z$,
the encoder must invert the generator at that $\z$, such that $E(G(\z)) = \z$.

\begin{restatable}{theorem}{lzeroautoenc}
  \label{thm:kld}
  The encoder and generator objective given an optimal discriminator
  $C(E,G) \defeq \max_D V(D, E, G)$
  can be rewritten as an $\ell_0$ autoencoder loss function
  \begin{align*}
  C(E, G)
  &=
  \expect{\x \sim p_\X}{\ldeltafunc{E(\x) \in \hOZ \land G(E(\x))=\x} \log f_{EG}(\x, E(\x))} + \\
  &\phantom{=} \; \; \;
  \expect{\z \sim p_\Z}{\ldeltafunc{G(\z) \in \hOX \land E(G(\z))=\z} \logp{1 - f_{EG}(G(\z), \z)}}
  \end{align*}
  with $\log f_{EG} \in (-\infty, 0)$ and $\logp{ 1 - f_{EG} } \in (-\infty, 0)$ $P_{E\X}$-almost and $P_{G\Z}$-almost everywhere.
\end{restatable}
\proof Given in Appendix \ref{sec:proof_aenc}.

Here the indicator function $\ldeltafunc{G(E(\x))=\x}$ in the first term is equivalent to an autoencoder with $\ell_0$ loss,
while the indicator $\ldeltafunc{E(G(\z))=\z}$ in the second term shows that the BiGAN encoder must invert the generator,
the desired property for feature learning.
The objective further encourages the functions $E(\x)$ and $G(\z)$ to produce valid outputs in the support of $P_\Z$ and $P_\X$ respectively.
Unlike regular autoencoders, the $\ell_0$ loss function does not make any assumptions about the structure or distribution of the data itself;
in fact, all the structural properties of BiGAN are learned as part of the discriminator.

\subsection{Learning}
\lblsec{learning}
In practice, as in the GAN framework~\citep{gan}, each BiGAN module
$D$, $G$, and $E$
is a parametric function
(with parameters
$\theta_D$, $\theta_G$, and $\theta_E$,
respectively).
As a whole, BiGAN can be optimized using alternating stochastic gradient steps.
In one iteration, the discriminator parameters $\theta_D$
are updated by taking one or more steps in the positive gradient direction $
\nabla_{\theta_D}
V(D, E, G)
$, then the encoder parameters $\theta_E$ and generator parameters $\theta_G$
are together updated by taking a step in the negative gradient direction $
-\nabla_{\theta_E,\theta_G}
V(D, E, G)
$.
In both cases, the expectation terms of $V(D, E, G)$ are estimated using mini-batches of $n$ samples $
\{\x^{(i)} \sim p_\X\}_{i=1}^n
$ and $
\{\z^{(i)} \sim p_\Z\}_{i=1}^n
$ drawn independently for each update step.

\citet{gan} found that
an objective in which the real and generated labels $Y$ are swapped
provides stronger gradient signal to $G$.
We similarly observed in BiGAN training that an ``inverse'' objective
provides stronger gradient signal to $G$ and $E$.
For efficiency, we also update all modules $D$, $G$, and $E$ simultaneously at each iteration,
rather than alternating between $D$ updates and $G$, $E$ updates.
See Appendix \ref{sec:learning_details} for details.

\subsection{Generalized BiGAN}
\label{sec:genbigan}
It is often useful to parametrize the output of the generator $G$ and encoder $E$ in a different, usually smaller, space $\OX'$ and $\OZ'$ rather than the original $\OX$ and $\OZ$.
For example, for visual feature learning, the images input to the encoder should be of similar resolution to images used in the evaluation.
On the other hand, generating high resolution images remains difficult for current generative models.
In this situation, the encoder may take higher resolution input while the generator output and discriminator input remain low resolution.

We generalize the BiGAN objective $V(D, G, E)$ (\ref{eq:vdge}) with functions $g_\X : \OX \mapsto \OX'$ and $g_\Z : \OZ \mapsto \OZ'$, and encoder $E: \OX \mapsto \OZ'$, generator $G: \OZ \mapsto \OX'$, and discriminator $D: \OX' \times \OZ' \mapsto [0, 1]$:
\begin{align*}
\mathbb{E}_{\x \sim p_\X}\big[
  \underbrace{
    \expect{\z' \sim p_E(\cdot | \x)}{
      \log D(g_\X(\x), \z')
    }
  }_{
    \log D(g_\X(\x), E(\x))
  }
\big]
+
\mathbb{E}_{\z \sim p_\Z}\big[
  \underbrace{
    \expect{\x' \sim p_G(\cdot | \z)}{
      \logp{1 - D(\x', g_\Z(\z))}
    }
  }_{
    \logp{1 - D(G(\z), g_\Z(\z))}
  }
\big]
\end{align*}
An identity $g_\X(\x)=\x$ and $g_\Z(\z)=\z$ (and $\OX'=\OX$, $\OZ'=\OZ$) yields the original objective.
For visual feature learning with higher resolution encoder inputs, $g_\X$ is an image resizing function that downsamples a high resolution image $\x \in \OX$ to a lower resolution image $\x' \in \OX'$, as output by the generator.  ($g_\Z$ is identity.)

In this case, the encoder and generator respectively induce probability measures $P_{E\X'}$ and $P_{G\Z'}$ over regions $R\subseteq\Omega'$ of the joint space $\Omega' \defeq \OX' \times \OZ'$, with $
P_{E\X'}(R) \defeq \smallint_{\OX} \smallint_{\OX'} \smallint_{\OZ'} p_{E\X}(\x, \z') \ldeltafunc{(\x', \z') \in R} \delta(g_\X(\x) - \x') \d\z' \d\x' \d\x
=
\smallint_{\OX} p_\X(\x) \ldeltafunc{(g_\X(\x), E(\x)) \in R} \d\x
$, and $P_{G\Z'}$ defined analogously.
For optimal $E$ and $G$, we can show $P_{E\X'} = P_{G\Z'}$: a generalization of Theorem~\ref{thm:optgen}.
When $E$ and $G$ are deterministic and optimal, Theorem~\ref{th:inverse} -- that $E$ and $G$ invert one another -- can also be generalized:
$\exists_{\z \in \hOZ} \{ E(\x) = g_\Z(\z) \,\land\, G(\z) = g_\X(\x) \}$ for $P_\X$-almost every $\x \in \OX$, and
$\exists_{\x \in \hOX} \{ E(\x) = g_\Z(\z) \,\land\, G(\z) = g_\X(\x) \}$ for $P_\Z$-almost every $\z \in \OZ$.

\section{Evaluation}
\lblsec{eval}

We evaluate the feature learning capabilities of BiGANs by first training them unsupervised as described in~\refsec{learning},
then transferring the encoder's learned feature representations for use in auxiliary supervised learning tasks.
To demonstrate that BiGANs are able to learn meaningful feature representations both on arbitrary data vectors,
where the model is agnostic to any underlying structure,
as well as very high-dimensional and complex distributions,
we evaluate on both permutation-invariant MNIST~\citep{mnist}
and on the high-resolution natural images of ImageNet~\citep{imagenet}.

In all experiments, each module $D$, $G$, and $E$ is a parametric deep (multi-layer) network.
The BiGAN discriminator $D(\x, \z)$ takes data $\x$ as its initial input, and at each linear layer thereafter,
the latent representation $\z$ is transformed using a learned linear transformation to the hidden layer dimension and added to the non-linearity input.

\subsection{Baseline methods}
\lblsec{baselines}
Besides the BiGAN framework presented above,
we considered alternative approaches to learning feature representations using different GAN variants.

\paragraph{Discriminator}
The discriminator $D$ in a standard GAN takes data samples $\x \sim p_\X$ as input,
making its learned intermediate representations natural candidates as feature representations for related tasks.
This alternative is appealing as it requires no additional machinery,
and is the approach used for unsupervised feature learning in~\citet{dcgan}.
On the other hand, it is not clear that the task of distinguishing between real and generated data requires or benefits from intermediate representations that are useful as semantic feature representations.
In fact, if $G$ successfully generates the true data distribution $p_\X(\x)$,
$D$ may ignore the input data entirely and predict $P(Y=1)=P(Y=1|\x)=\frac{1}{2}$ unconditionally, not learning any meaningful intermediate representations.

\paragraph{Latent regressor}
\lblsec{baselineenc}
We consider an alternative encoder training by minimizing a reconstruction loss $
\mathcal{L}(\z, E(G(\z)))$, after or jointly during a regular GAN training, called latent regressor or joint latent regressor respectively.
We use a sigmoid cross entropy loss $\mathcal{L}$ as it naturally maps to a uniformly distributed output space.
Intuitively, a drawback of this approach is that,
unlike the encoder in a BiGAN,
the latent regressor encoder $E$ is trained only on generated samples $G(\z)$,
and never ``sees'' real data $\x \sim p_\X$.
While this may not be an issue in the theoretical optimum where $
p_G(\x)
=
p_\X(\x)
$ exactly
-- i.e., $G$ perfectly generates the data distribution $p_\X$ --
in practice, for highly complex data distributions $p_\X$,
such as the distribution of natural images,
the generator will almost never achieve this perfect result.
The fact that the real data $\x$ are never input to this type of encoder
limits its utility as a feature representation for related tasks, as shown later in this section.

\subsection{Permutation-invariant MNIST}
\lblsec{evalmnist}
We first present results on permutation-invariant MNIST~\citep{mnist}.
In the permutation-invariant setting,
each $28\times28$ digit image must be treated as an unstructured $784$D vector~\citep{maxout}.
In our case, this condition is met by designing each module as a multi-layer perceptron (MLP),
agnostic to the underlying spatial structure in the data
(as opposed to a convnet, for example).
See Appendix \ref{sec:mnistarch} for more architectural and training details.
We set the latent distribution $p_\Z = \left[\mathrm{U}(-1, 1)\right]^{50}$ -- a $50$D continuous uniform distribution.

\reftbl{mnist} compares the encoding learned by a BiGAN-trained encoder $E$
with the baselines described in \refsec{baselines},
as well as autoencoders~\citep{hinton-autoencoders}
trained directly to minimize either
$\ell_2$ or $\ell_1$ reconstruction error.
The same architecture and optimization algorithm is used across all methods.
All methods, including BiGAN, perform at roughly the same level.
This result is not overly surprising given the relative simplicity of MNIST digits.
For example, digits generated by $G$ in a GAN nearly perfectly match the data distribution (qualitatively),
making the latent regressor (LR) baseline method a reasonable choice, as argued in \refsec{baselineenc}.
Qualitative results are presented in \reffig{mnistqual}.

\input{table_mnist}
\input{qual_mnist}

\subsection{ImageNet}
\lblsec{evalimagenet}
Next, we present results from training BiGANs on ImageNet LSVRC~\citep{imagenet},
a large-scale database of natural images.
GANs trained on ImageNet cannot perfectly reconstruct the data, but often capture some interesting aspects.
Here, each of $D$, $G$, and $E$ is a convnet.
In all experiments, the encoder $E$ architecture follows AlexNet~\citep{supervision}
through the fifth and last convolution layer (\textit{conv5}).
We also experiment with an AlexNet-based discriminator $D$ as a baseline feature learning approach.
We set the latent distribution $p_\Z = \left[\mathrm{U}(-1, 1)\right]^{200}$
-- a $200$D continuous uniform distribution.
Additionally, we experiment with higher resolution encoder input images -- $112\times112$ rather than the $64\times64$ used elsewhere -- using the generalization described in Section~\ref{sec:genbigan}.
See Appendix \ref{sec:imagenetarch} for more architectural and training details.

\input{imagenet_filters}

\paragraph{Qualitative results}
The convolutional filters learned by each of the three modules are shown in \reffig{filters}.
We see that the filters learned by the encoder $E$ have clear Gabor-like structure,
similar to those originally reported for the fully supervised AlexNet model~\citep{supervision}.
The filters also have similar ``grouping'' structure where one half (the bottom half, in this case) is more color sensitive,
and the other half is more edge sensitive.
(This separation of the filters occurs due to the AlexNet architecture
maintaining two separate filter paths for computational efficiency.)

\input{qual_imagenet}
In \reffig{imagenetqual} we present sample generations $G(\z)$,
as well as real data samples $\x$ and their BiGAN reconstructions $G(E(\x))$.
The reconstructions, while certainly imperfect,
demonstrate empirically that the BiGAN encoder $E$ and generator $G$ learn approximate inverse mappings,
as shown theoretically in Theorem \ref{th:inverse}.
In Appendix \ref{sec:nn_imagenet}, we present nearest neighbors in the BiGAN learned feature space.

\paragraph{ImageNet classification}

Following~\citet{jigsaw}, we evaluate by freezing the first $N$ layers of our pretrained network and randomly reinitializing and training the remainder fully supervised for ImageNet classification.
Results are reported in \reftbl{imagenetclass}.
\input{table_imagenet_class}

\paragraph{VOC classification, detection, and segmentation}
We evaluate the transferability of BiGAN representations to the PASCAL VOC~\citep{pascal} computer vision benchmark tasks,
including classification, object detection, and semantic segmentation.
The classification task involves simple binary prediction of presence or absence in a given image for each of 20 object categories.
The object detection and semantic segmentation tasks go a step further by requiring the objects to be localized,
with semantic segmentation requiring this at the finest scale: pixelwise prediction of object identity.
For detection, the pretrained model is used as the initialization for \textit{Fast R-CNN}~\citep{fastrcnn} (FRCN) training;
and for semantic segmentation, the model is used as the initialization for \textit{Fully Convolutional Network}~\citep{fcn} (FCN) training,
in each case replacing the \textit{AlexNet}~\citep{supervision} model trained fully supervised for ImageNet classification.
We report results on each of these tasks in \reftbl{vocclass},
comparing BiGANs with contemporary approaches to unsupervised~\citep{datadep}
and self-supervised~\citep{carl,pulkit,xiaolong,deepak}
feature learning in the visual domain,
as well as the baselines discussed in \refsec{baselines}.
\input{table_voc_all}

\subsection{Discussion}
Despite making no assumptions about the underlying structure of the data,
the BiGAN unsupervised feature learning framework offers a representation competitive with existing self-supervised and even weakly supervised feature learning approaches for visual feature learning,
while still being a purely generative model with the ability to sample data $\x$ and predict latent representation $\z$.
Furthermore, BiGANs outperform the discriminator ($D$) and latent regressor (LR) baselines discussed in \refsec{baselines},
confirming our intuition that these approaches may not perform well in the regime of highly complex data distributions such as that of natural images.
The version in which the encoder takes a higher resolution image than output by the generator (\textit{BiGAN $112\times112$ $E$}) performs better still,
and this strategy is not possible under the LR and $D$ baselines as each of those modules take generator outputs as their input.

Although existing self-supervised approaches have shown impressive performance and thus far tended to outshine purely unsupervised approaches
in the complex domain of high-resolution images,
purely unsupervised approaches to feature learning or pre-training have several potential benefits.

BiGAN and other unsupervised learning approaches are agnostic to the domain of the data.
The self-supervised approaches are specific to the visual domain,
in some cases requiring weak supervision from video unavailable in images alone.
For example, the methods are not applicable in the permutation-invariant MNIST setting explored in Section~\ref{sec:evalmnist},
as the data are treated as flat vectors rather than 2D images.

Furthermore, BiGAN and other unsupervised approaches needn't suffer from domain shift between the pre-training task and the transfer task,
unlike self-supervised methods in which some aspect of the data is normally removed or corrupted in order to create a non-trivial prediction task.
In the context prediction task~\citep{carl}, the network sees only small image patches --
the global image structure is unobserved.
In the context encoder or inpainting task~\citep{deepak},
each image is corrupted by removing large areas to be filled in by the prediction network,
creating inputs with dramatically different appearance
from the uncorrupted natural images seen in the transfer tasks.

Other approaches~\citep{pulkit,xiaolong} rely on auxiliary information
unavailable in the static image domain,
such as video, egomotion, or tracking.
Unlike BiGAN, such approaches cannot learn feature representations from unlabeled static images.

We finally note that the results presented here constitute only a preliminary exploration of the space of model architectures possible under the BiGAN framework,
and we expect results to improve significantly with advancements in generative image models and discriminative convolutional networks alike.

\subsubsection*{Acknowledgments}
The authors thank Evan Shelhamer, Jonathan Long, and other Berkeley Vision labmates for helpful discussions throughout this work.
This work was supported by DARPA, AFRL, DoD MURI award N000141110688, NSF awards IIS-1427425 and IIS-1212798, and the Berkeley Artificial Intelligence Research laboratory.
The GPUs used for this work were donated by NVIDIA.

\newpage
\bibliography{paper}
\bibliographystyle{paper}

\newpage
\begin{appendices}
\section{Additional proofs}
\subsection{Proof of Proposition~\ref{eq:optdiscrim} (optimal discriminator)}
\lblsec{optdiscrim}
\optdiscrim*
\proof
For measures $P$ and $Q$ on space $\Omega$, with $P$ absolutely continuous with respect to $Q$,
the RN derivative $
f_{PQ} \defeq \tfrac{\d P}{\d Q}
$ exists, and we have
\begin{equation}
\expect{\x \sim P}{
  g(\x)
}
=
\smallint_{\Omega}
  g \d P
=
\smallint_{\Omega}
  g
  \tfrac{\d P}{\d Q}
  \d Q
=
\smallint_{\Omega}
  g
  f_{PQ}
  \d Q
=
\expect{\x \sim Q}{
  f_{PQ}(\x)
  g(\x)
}.
\label{eq:egpq}
\end{equation}
Let the probability measure $
P_{EG} \defeq
\tfrac{P_{E\X} + P_{G\Z}}{2}
$ denote the average of measures $P_{E\X}$ and $P_{G\Z}$.
Both $P_{E\X}$ and $P_{G\Z}$ are each absolutely continuous with respect to $P_{EG}$.
Hence the RN derivatives $
f_{EG} \defeq \tfrac{\d P_{E\X}}{\d (P_{E\X} + P_{G\Z})}
= \tfrac{1}{2} \tfrac{\d P_{E\X}}{\d P_{EG}}
$ and $
f_{GE} \defeq \tfrac{\d P_{G\Z}}{\d (P_{E\X} + P_{G\Z})}
= \tfrac{1}{2} \tfrac{\d P_{G\Z}}{\d P_{EG}}
$ exist and sum to $1$:
\begin{equation}
f_{EG} + f_{GE}
=
\tfrac{\d P_{E\X}}{\d (P_{E\X} + P_{G\Z})} +
\tfrac{\d P_{G\Z}}{\d (P_{E\X} + P_{G\Z})}
= \tfrac{\d (P_{E\X} + P_{G\Z})}{\d (P_{E\X} + P_{G\Z})}
= 1.
\label{fegpfge}
\end{equation}
We use~\eqref{eq:egpq} and \eqref{fegpfge} to rewrite the objective $V$ \eqref{eq:vdge} as a single expectation under measure $P_{EG}$:
\begin{align}
\nonumber
V(D, E, G)
&=
\expect{(\x, \z) \sim P_{E\X}}{
  \log D(\x, \z)
}
+
\expect{(\x, \z) \sim P_{G\Z}}{
  \logp{ 1 - D(\x, \z) }
}
\\ \nonumber
&=
\mathbb{E}_{(\x, \z) \sim P_{EG}}[
  \underbrace{2 f_{EG}}_{
    \tfrac{\d P_{E\X}}{\d P_{EG}}
  }(\x, \z)
  \log D(\x, \z)
]
+
\mathbb{E}_{(\x, \z) \sim P_{EG}}[
  \underbrace{2 f_{GE}}_{
    \tfrac{\d P_{G\Z}}{\d P_{EG}}
  }(\x, \z)
  \logp{ 1 - D(\x, \z) }
]
\\ \nonumber
&=
2 \,
\expect{(\x, \z) \sim P_{EG}}{
  f_{EG}(\x, \z)
  \log D(\x, \z)
+
  f_{GE}(\x, \z)
  \logp{ 1 - D(\x, \z) }
}
\\ \nonumber
&=
2 \,
\expect{(\x, \z) \sim P_{EG}}{
  f_{EG}(\x, \z)
  \log D(\x, \z)
+
  (1-f_{EG}(\x, \z))
  \logp{ 1 - D(\x, \z) }
}.
\end{align}
Note that $
\argmax_y
\left\{
a \log y + (1-a) \log (1-y)
\right\}
=
a
$ for any $
a \in [0,1]
$.
Thus, $D^*_{EG}=f_{EG}$.
$\square$

\subsection{Proof of Proposition~\ref{cegjsd} (encoder and generator objective)}
\lblsec{cegjsd}
\cegjsd*
\proof
Using Proposition~\ref{eq:optdiscrim} along with~\eqref{fegpfge} ($
1 - D^*_{EG}
=
1 - f_{EG}
=
f_{GE}$) we rewrite the objective
\begin{align}
\nonumber
C(E, G)
&=
{\begingroup\textstyle \max\endgroup}_D
V(D, E, G)
=
V(D^*_{EG}, E, G)
\\
\nonumber
&=
\expect{(\x, \z) \sim P_{E\X}}{
  \log D^*_{EG}(\x, \z)
}
+
\expect{(\x, \z) \sim P_{G\Z}}{
  \logp{ 1 - D^*_{EG}(\x, \z) }
}
\\
\nonumber
&=
\expect{(\x, \z) \sim P_{E\X}}{
  \log f_{EG}(\x, \z)
}
+
\expect{(\x, \z) \sim P_{G\Z}}{
  \log f_{GE}(\x, \z)
}
\\
\nonumber
&=
\expect{(\x, \z) \sim P_{E\X}}{
  \logp{ 2 f_{EG}(\x, \z) }
}
+
\expect{(\x, \z) \sim P_{G\Z}}{
  \logp{ 2 f_{GE}(\x, \z) }
}
- \log 4
\\ \nonumber
&=
\kld{P_{E\X}}{P_{EG}}
+
\kld{P_{G\Z}}{P_{EG}}
-
\log 4
\\
\nonumber
&=
\kld{P_{E\X}}{\tfrac{P_{E\X}+P_{G\Z}}{2}}
+
\kld{P_{G\Z}}{\tfrac{P_{E\X}+P_{G\Z}}{2}}
-
\log 4
\\
\nonumber
&=
2 \, \jsd{P_{E\X}}{P_{G\Z}}
-
\log 4.
\; \square
\end{align}

\subsection{Measure definitions for deterministic $E$ and $G$}
\label{sec:deteg}
While Theorem~\ref{thm:optgen} and Propositions~\ref{eq:optdiscrim} and~\ref{cegjsd} hold for any encoder $p_E(\z | \x)$ and generator $p_G(\x | \z)$, stochastic or deterministic,
Theorems~\ref{th:inverse} and~\ref{thm:kld} assume the encoder $E$ and generator $G$ are deterministic functions;
i.e., with conditionals $
p_E(\z | \x) = \delta(\z - E(\x))
$ and $
p_G(\x | \z) = \delta(\x - G(\z))
$ defined as $\delta$ functions.

For use in the proofs of those theorems, we simplify the definitions of measures $P_{E\X}$ and $P_{G\Z}$ given in Section~\ref{sec:defs} for the case of deterministic functions $E$ and $G$ below:
\begin{align*}
  P_{E\X}(R)
  &= \smallint_{\OX} p_\X(\x) \smallint_{\OZ} p_E(\z | \x) \ldeltafunc{(\x, \z) \in R} \d\z \d\x \\
  &= \smallint_{\OX} p_\X(\x) \left( \smallint_{\OZ} \delta(\z - E(\x)) \ldeltafunc{(\x, \z) \in R} \d\z \right) \d\x \\
  &= \smallint_{\OX} p_\X(\x) \ldeltafunc{(\x, E(\x)) \in R} \d\x \\
  P_{G\Z}(R)
  &= \smallint_{\OZ} p_\Z(\z) \smallint_{\OX} p_G(\x | \z) \ldeltafunc{(\x, \z) \in R} \d\x \d\z \\
  &= \smallint_{\OZ} p_\Z(\z) \left( \smallint_{\OX} \delta(\x - G(\z)) \ldeltafunc{(\x, \z) \in R} \d\x \right) \d\z \\
  &= \smallint_{\OZ} p_\Z(\z) \ldeltafunc{(G(\z), \z) \in R} \d\z
\end{align*}

\subsection{Proof of Theorem~\ref{th:inverse} (optimal generator and encoder are inverses)}
\label{sec:geinv}
\thinv*
\proof
Let $R^0_\X \defeq \{ \x \in \OX : \x \ne G(E(\x)) \}$ be the region of $\OX$ in which the inversion property $\x = G(E(\x))$ does \textit{not} hold.
We will show that, for optimal $E$ and $G$, $R^0_\X$ has measure zero under $P_\X$ (i.e., $P_\X(R^0_\X)=0$) and therefore $\x = G(E(\x))$ holds $P_\X$-almost everywhere.

Let $R^0 \defeq \{ (\x, \z) \in \Omega : \z = E(\x) \,\land\, \x \in R^0_\X \}$ be the region of $\Omega$ such that
$(\x, E(\x)) \in R^0$ if and only if $\x \in R^0_\X$.
We'll use the definitions of $P_{E\X}$ and $P_{G\Z}$ for deterministic $E$ and $G$ (Appendix~\ref{sec:deteg}),
and the fact that $P_{E\X} = P_{G\Z}$ for optimal $E$ and $G$ (Theorem~\ref{thm:optgen}).
\begin{align*}
  P_\X(R^0_\X)
  &= \smallint_{\OX} p_\X(\x) \ldeltafunc{\x \in R^0_\X} \d\x \\
  &= \smallint_{\OX} p_\X(\x) \ldeltafunc{(\x, E(\x)) \in R^0} \d\x \\
  &= P_{E\X}(R^0) \\
  &= P_{G\Z}(R^0) \\
  &= \smallint_{\OZ} p_\Z(\z) \ldeltafunc{(G(\z), \z) \in R^0} \d\z \\
  &= \smallint_{\OZ} p_\Z(\z) \ldeltafunc{\z = E(G(\z)) \,\land\, G(\z) \in R^0_\X} \d\z \\
  &= \smallint_{\OZ} p_\Z(\z) \underbrace{\ldeltafunc{\z = E(G(\z)) \,\land\, G(\z) \ne G(E(G(\z)))}}_{=0\text{ for any }\z\text{, as } \z=E(G(\z)) \implies G(\z) = G(E(G(\z)))} \d\z \\
  &= 0.
\end{align*}
Hence region $R^0_\X$ has measure zero ($P_\X(R^0_\X) = 0$), and the inversion property $\x=G(E(\x))$ holds $P_\X$-almost everywhere.

An analogous argument shows that
$R^0_\Z \defeq \{ \z \in \OZ : \z \ne E(G(\z)) \}$ has measure zero on $P_\Z$ (i.e., $P_\Z(R^0_\Z)=0$) and therefore
$\z=E(G(\z))$ holds $P_\Z$-almost everywhere.
$\square$

\subsection{Proof of Theorem~\ref{thm:kld} (relationship to autoencoders)}
\lblsec{proof_aenc}
As shown in Proposition~\ref{cegjsd} (\refsec{bigan}), the BiGAN objective is equivalent to the Jensen-Shannon divergence between $P_{E\X}$ and $P_{G\Z}$.
We now go a step further and show that this Jensen-Shannon divergence is closely related to a standard autoencoder loss.
Omitting the $\tfrac{1}{2}$ scale factor, a KL divergence term of the Jensen-Shannon divergence is given as
\begin{align}
  \kld{P_{E\X}}{\tfrac{P_{E\X}+P_{G\Z}}{2}}
  \nonumber
  &= \log 2 + \int_{\Omega} \log \frac{\d P_{E\X}}{\d(P_{E\X} + P_{G\Z})} \d P_{E\X} \\
  &= \log 2 + \int_{\Omega} \log f \d P_{E\X}
  ,
\label{autoenckldiv}
\end{align}
where we abbreviate as $f$ the Radon-Nikodym derivative $
f_{EG}
\defeq
\frac{\d P_{E\X}}{\d \left(P_{E\X} + P_{G\Z}\right)}
\in [0, 1]
$ defined in Proposition~\ref{eq:optdiscrim} for most of this proof.

We'll make use of the definitions of $P_{E\X}$ and $P_{G\Z}$ for deterministic $E$ and $G$ found in Appendix~\ref{sec:deteg}.
The integral term of the KL divergence expression given in~\eqref{autoenckldiv} over a particular region $R \subseteq \Omega$ will be denoted by
$$F(R) \defeq \int_R \log \frac{\d P_{E\X}}{\d\left(P_{E\X} + P_{G\Z}\right)} \d P_{E\X}
=
\int_R \log f \d P_{E\X}
.$$
Next we will show that $f > 0$ holds $P_{E\X}$-almost everywhere, and hence $F$ is always well defined and finite.
We then show that $F$ is equivalent to an autoencoder-like reconstruction loss function.

\begin{proposition}
$f > 0$ $P_{E\X}$-almost everywhere. \label{prop:fgt0}
\end{proposition}
\proof
Let $R^{f=0} \defeq \left\{ (\x, \z) \in \Omega : f(\x, \z) = 0 \right\}$ be the region of $\Omega$ in which $f = 0$.
Using the definition of the Radon-Nikodym derivative $f$, the measure 
$
P_{E\X}(R^{f=0})
= \smallint_{R^{f=0}} f \d (P_{E\X} + P_{G\Z})
= \smallint_{R^{f=0}} 0 \d (P_{E\X} + P_{G\Z})
= 0
$
is zero. Hence $f > 0$ $P_{E\X}$-almost everywhere.
$\square$

Proposition \ref{prop:fgt0} ensures
that $\log f$ is defined $P_{E\X}$-almost everywhere, and $F(R)$ is well-defined.
Next we will show that $F(R)$ mimics an autoencoder with $\ell_0$ loss, meaning $F$ is zero for any region in which $G(E(\x))\ne \x$, and non-zero otherwise.

\begin{proposition}
\label{prop:pgzsupport}
The KL divergence $F$ outside the support of $P_{G\Z}$ is zero: $F(\Omega \setminus \mathrm{supp}(P_{G\Z})) = 0$.
\end{proposition}
We'll first show that in region $R_S \defeq \Omega \setminus \mathrm{supp}(P_{G\Z})$, we have $f = 1$ $P_{E\X}$-almost everywhere.
Let $R^{f < 1} \defeq \left\{ (\x, \z) \in R_S : f(\x, \z) < 1 \right\}$ be the region of $R_S$ in which $f < 1$.
Let's assume that $P_{E\X}(R^{f<1}) > 0$ has non-zero measure.
Then, using the definition of the Radon-Nikodym derivative,
\begin{align*}
 P_{E\X}(R^{f<1}) &= \smallint_{R^{f<1}} f \d (P_{E\X} + P_{G\Z}) = \smallint_{R^{f<1}} \underbrace{f}_{\le \varepsilon < 1} \d P_{E\X} + \underbrace{\smallint_{R^{f<1}} f \d P_{G\Z}}_{0} \le \varepsilon P_{E\X}(R^{f<1})\\
 &< P_{E\X}(R^{f<1}),
\end{align*}
where $\varepsilon$ is a constant smaller than $1$.
But $P_{E\X}(R^{f<1}) < P_{E\X}(R^{f<1})$ is a contradiction;
hence $P_{E\X}(R^{f<1}) = 0$ and $f = 1$ $P_{E\X}$-almost everywhere in $R_S$,
implying $\log f = 0$ $P_{E\X}$-almost everywhere in $R_S$.
Hence $F(R_S)=0$. $\square$

By definition, $F(\Omega \setminus \mathrm{supp}(P_{E\X})) = 0$ is also zero.
The only region where $F$ might be non-zero is $R^{1} \defeq \mathrm{supp}(P_{E\X}) \cap \mathrm{supp}(P_{G\Z})$.

\begin{proposition}
\label{prop:flt1}
$f < 1$ $P_{E\X}$-almost everywhere in $R^1$.
\end{proposition}
Let $R^{f=1} \defeq \left\{ (\x, \z) \in R^1 : f(\x, \z) = 1 \right\}$ be the region in which $f=1$.
Let's assume the set $R^{f=1} \ne \emptyset$ is not empty.
By definition of the support\footnote{We use the definition $U\cap C\neq\emptyset \implies \mu (U\cap C)>0$ here.},
$P_{E\X}(R^{f=1}) > 0$ and $P_{G\Z}(R^{f=1})>0$.
The Radon-Nikodym derivative on $R^{f=1}$ is then given by
\begin{align*}
 P_{E\X}(R^{f=1}) = \smallint_{R^{f=1}} f \d (P_{E\X} + P_{G\Z}) = \smallint_{R^{f=1}} 1 \d (P_{E\X} + P_{G\Z}) = P_{E\X}(R^{f=1}) + P_{G\Z}(R^{f=1}),
\end{align*}
which implies $P_{G\Z}(R^{f=1}) = 0$ and contradicts the definition of support. Hence $R^{f=1} = \emptyset$ and $f<1$ $P_{E\X}$-almost everywhere on $R^1$, implying $\log f < 0$ $P_{E\X}$-almost everywhere. $\square$

\lzeroautoenc*
\proof
Proposition~\ref{prop:pgzsupport}
($F(\Omega \setminus \mathrm{supp}(P_{G\Z})) = 0$)
and 
$ F(\Omega \setminus \mathrm{supp}(P_{E\X})) = 0 $
imply that $
R^1 \defeq
\mathrm{supp}(P_{E\X}) \cap \mathrm{supp}(P_{G\Z})
$ is the only region of $\Omega$ where $F$ may be non-zero; hence $F(\Omega) = F(R^1)$.
Note that
\begin{align*}
\supp(P_{E\X}) &=
\{ (\x,E(\x)) : \x \in \hOX \}
\\
\supp(P_{G\Z}) &=
\{ (G(\z),\z) : \z \in \hOZ \} \\
\implies R^1 \defeq
\supp(P_{E\X}) \cap \supp(P_{G\Z}) &=
\{
(\x,\z) :
E(\x) = \z \land
\x \in \hOX \land
G(\z) = \x \land
\z \in \hOZ
\}
\end{align*}
So a point $(\x, E(\x))$ is in $R^1$ if $\x \in \hOX$, $E(\x) \in \hOZ$, and $G(E(\x))=\x$.
(We can omit the $\x \in \hOX$ condition from inside an expectation over $P_\X$, as $P_\X$-almost all $\x \notin \hOX$ have 0 probability.)
Therefore,
\begin{align*}
  \kld{P_{E\X}}{\tfrac{P_{E\X}+P_{G\Z}}{2}}
  - \log 2
  &= F(\Omega) = F(R^1) \\
  &= \smallint_{R^1} \log f(\x, \z) \d P_{E\X} \\
  &= \smallint_{\Omega} \ldeltafunc{(\x, \z) \in R^1} \log f(\x, \z) \d P_{E\X} \\
  &= \expect{(\x, \z) \sim P_{E\X}}{\ldeltafunc{(\x, \z) \in R^1} \log f(\x, \z)} \\
  &= \expect{\x \sim p_\X}{\ldeltafunc{(\x, E(\x)) \in R^1} \log f(\x, E(\x))} \\
  &= \expect{\x \sim p_\X}{
    \ldeltafunc{E(\x) \in \hOZ \land G(E(\x))=\x} \log f(\x, E(\x))
  }.
\end{align*}
Finally, with Propositions \ref{prop:fgt0} and \ref{prop:flt1},
we have $f \in (0, 1)$ $P_{E\X}$-almost everywhere in $R^1$,
and therefore
$\log f \in (-\infty, 0)$, taking a finite and strictly negative value $P_{E\X}$-almost everywhere.

An analogous argument
(along with the fact that $f_{EG} + f_{GE} = 1$) lets us rewrite the other KL divergence term
\begin{align*}
\kld{P_{G\Z}}{\tfrac{P_{E\X}+P_{G\Z}}{2}}
- \log 2
&= \expect{\z \sim p_\Z}{\ldeltafunc{G(\z) \in \hOX \land E(G(\z))=\z} \log f_{GE}(G(\z), \z)} \\
&= \expect{\z \sim p_\Z}{\ldeltafunc{G(\z) \in \hOX \land E(G(\z))=\z} \logp{ 1 - f_{EG}(G(\z), \z) }}
\end{align*}
The Jensen-Shannon divergence is the mean of these two KL divergences, giving $C(E, G)$:
\begin{align*}
  C(E, G)
  &= 2 \, \jsd{P_{E\X}}{P_{G\Z}} - \log 4
  \\
  &=
  \kld{P_{E\X}}{\tfrac{P_{E\X}+P_{G\Z}}{2}}
  +
  \kld{P_{G\Z}}{\tfrac{P_{E\X}+P_{G\Z}}{2}}
  - \log 4
  \\
  &=
  \expect{\x \sim p_\X}{\ldeltafunc{E(\x) \in \hOZ \land G(E(\x))=\x} \log f_{EG}(\x, E(\x))} + \\
  &\phantom{=} \; \; \;
  \expect{\z \sim p_\Z}{\ldeltafunc{G(\z) \in \hOX \land E(G(\z))=\z} \logp{ 1 - f_{EG}(G(\z), \z) }}
  \square
\end{align*}

\section{Learning details}
\lblsec{learning_details}
In this section we provide additional details on the BiGAN learning protocol summarized in \refsec{learning}.
\citet{gan} found for GAN training that
an objective in which the real and generated labels $Y$ are swapped
provides stronger gradient signal to $G$.
We similarly observed in BiGAN training that an ``inverse'' objective $\Lambda$
(with the same fixed point characteristics as $V$)
provides stronger gradient signal to $G$ and $E$, where
\begin{align*}
\Lambda(D, G, E)
=
\mathbb{E}_{\x \sim p_\X}\big[
  \underbrace{
    \expect{\z \sim p_E(\cdot | \x)}{
      \logp{1 - D(\x, \z)}
    }
  }_{
    \logp{1 - D(\x, E(\x))}
  }
\big]
+
\mathbb{E}_{\z \sim p_\Z}\big[
  \underbrace{
    \expect{\x \sim p_G(\cdot | \z)}{
      \log D(\x, \z)
    }
  }_{
    \log D(G(\z), \z)
  }
\big].
\end{align*}
In practice, $\theta_G$ and $\theta_E$ are updated by moving in the
positive gradient direction of this inverse objective $
\nabla_{\theta_E,\theta_G}
\Lambda
$,
rather than the negative gradient direction of the original objective.

We also observed that learning behaved similarly when all parameters
$\theta_D$, $\theta_G$, $\theta_E$
were updated simultaneously at each iteration rather than alternating between
$\theta_D$ updates and
$\theta_G, \theta_E$ updates,
so we took the simultaneous updating (non-alternating) approach for computational efficiency.
(For standard GAN training,
simultaneous updates of $\theta_D$, $\theta_G$ performed similarly well,
so our standard GAN experiments also follow this protocol.)

\section{Model and training details}
In the following sections we present additional details on the models and training protocols used in the
permutation-invariant MNIST
and
ImageNet
evaluations presented in~\refsec{eval}.

\paragraph{Optimization}
For unsupervised training of BiGANs and baseline methods,
we use the Adam optimizer~\citep{adam} to compute parameter updates,
following the hyperparameters
(initial step size $
\alpha = 2\times10^{-4}
$, momentum $
\beta_1 = 0.5
$ and $
\beta_2 = 0.999
$)
used by~\citet{dcgan}.
The step size $\alpha$ is decayed exponentially to $
\alpha = 2\times10^{-6}
$ starting halfway through training.
The mini-batch size is 128.
$\ell_2$ weight decay of $
2.5\times10^{-5}
$
is applied to all multiplicative weights in linear layers
(but not to the learned bias $\beta$ or scale $\gamma$ parameters applied after batch normalization).
Weights are initialized from
a zero-mean normal distribution with a standard deviation of $0.02$,
with one notable exception:
BiGAN discriminator weights that directly multiply $\z$ inputs
to be added to spatial convolution outputs have initializations scaled by the convolution kernel size --
e.g., for a $5\times5$ kernel, weights are initialized with
a standard deviation of $0.5$,
$25$ times the standard initialization.

\paragraph{Software \& hardware}
We implement BiGANs and baseline feature learning methods using the \textit{Theano}~\citep{theano} framework,
based on the convolutional GAN implementation provided by~\citet{dcgan}.
ImageNet transfer learning experiments (\refsec{evalimagenet}) use the \textit{Caffe}~\citep{caffe} framework,
per the Fast R-CNN~\citep{fastrcnn} and FCN~\citep{fcn} reference implementations.
Most computation is performed on an NVIDIA Titan X or Tesla K40 GPU.

\subsection{Permutation-invariant MNIST}
\lblsec{mnistarch}
In all permutation-invariant MNIST experiments (\refsec{evalmnist}),
$D$, $G$, and $E$ each consist of two hidden layers with 1024 units.
The first hidden layer is followed by a non-linearity;
the second is followed by (parameter-free) batch normalization~\citep{batchnorm} and a non-linearity.
The second hidden layer in each case is the input to a linear prediction layer of the appropriate size.
In $D$ and $E$, a leaky ReLU~\citep{leakyrelu} non-linearity with a ``leak'' of 0.2 is used;
in $G$, a standard ReLU non-linearity is used.
All models are trained for 400 epochs.

\subsection{ImageNet}
\lblsec{imagenetarch}
In all ImageNet experiments (\refsec{evalimagenet}), the encoder $E$ architecture follows AlexNet~\citep{supervision}
through the fifth and last convolution layer (\textit{conv5}),
with local response normalization (LRN) layers removed and batch normalization~\citep{batchnorm}
(including the learned scaling and bias)
with leaky ReLU non-linearity applied to the output of each convolution at unsupervised training time.
(For supervised evaluation, batch normalization is not used, and the pre-trained scale and bias is merged into the preceding convolution's weights and bias.)

In most experiments, both the discriminator $D$ and generator $G$ architecture
are those used by~\citet{dcgan},
consisting of a series of four $5\times5$ convolutions
(or ``deconvolutions'' -- fractionally-strided convolutions -- for the generator $G$)
applied with 2 pixel stride,
each followed by batch normalization and rectified non-linearity.

The sole exception is our discriminator baseline feature learning experiment,
in which we let the discriminator $D$ be the AlexNet variant described above.
Generally, using AlexNet (or similar convnet architecture) as the discriminator $D$
is detrimental to the visual fidelity of the resulting generated images,
likely due to the relatively large convolutional filter kernel size applied to the input image,
as well as the max-pooling layers, which explicitly discard information in the input.
However, for fair comparison of the discriminator's feature learning abilities with those of BiGANs,
we use the same architecture as used in the BiGAN encoder.

\paragraph{Preprocessing}
To produce a data sample $\x$, we first sample an image from the database,
and resize it proportionally such that its shorter edge has a length of 72 pixels.
Then, a $64\times64$ crop is randomly selected from the resized image.
The crop is flipped horizontally with probability $\frac{1}{2}$.
Finally, the crop is scaled to $[-1, 1]$, giving the sample $\x$.

\paragraph{Timing}
A single epoch (one training pass over the 1.2 million images) of BiGAN training takes roughly 40 minutes on a Titan X GPU.
Models are trained for 100 epochs, for a total training time of under 3 days.

\paragraph{Nearest neighbors}
In \reffig{imagenetneighbors} we present nearest neighbors in the feature space of the BiGAN encoder $E$ learned in unsupervised ImageNet training.
\label{sec:nn_imagenet}
\input{nn_imagenet}

\end{appendices}
\end{document}

%% file: concept.tex
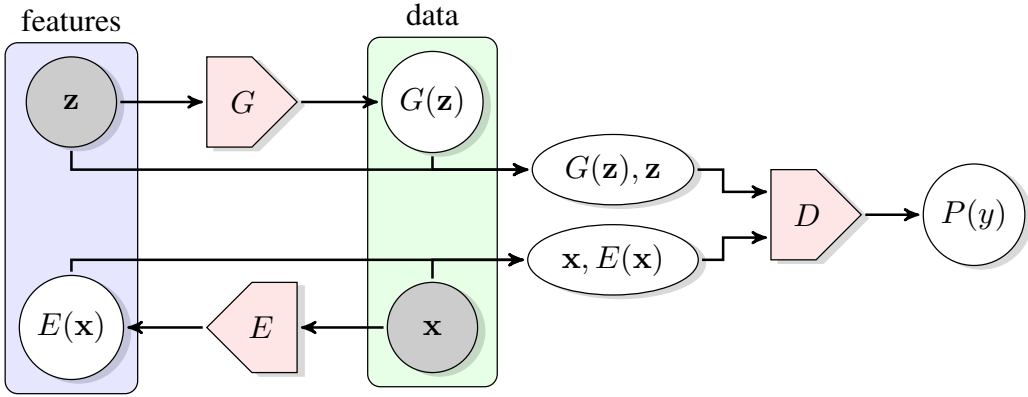
\begin{figure}[t]
\tikzstyle{shadow} = [copy shadow={opacity=0.15, shadow xshift=0.5ex,shadow yshift=-0.25ex,fill=black}]
\tikzstyle{all_nodes} = [minimum width=1cm, minimum height=0.5cm,text centered, draw=black,fill=white,shadow]
\tikzstyle{observed_var} = [circle, all_nodes, fill=black!20]
\tikzstyle{latent_var} = [circle, all_nodes]
\tikzstyle{group_var} = [ellipse, all_nodes, inner sep=0.1cm]
\tikzstyle{func} = [signal, signal to=east, all_nodes,minimum height=1cm,fill=red!10]
\tikzstyle{arrow}= [->,thick,shorten <=1pt,shorten >=1pt, >=stealth']
\tikzstyle{box}= [rectangle, rounded corners, draw=black, inner sep=0.15cm]
\tikzset{
  -|-/.style={
    to path={
      (\tikztostart) -| ($(\tikztostart)!#1!(\tikztotarget)$) |- (\tikztotarget)
      \tikztonodes
    }
  },
  -|-/.default=0.7,
}
\centering
\scalebox{1.2}{
 \begin{tikzpicture}[node distance=2cm]
  \node (z) [observed_var] at (-2,1.25) {$\z$};
  \node (G) [func] at (-0.25,1.25) {};
  \node (Gtext) at (-0.1,1.25) {$G$};
  \node (Gz) [latent_var] at (2,1.25) {$G(\z)$};
  \node (x) [observed_var] at (2,-1.25) {$\x$};
  \node (E) [func, shape border rotate=180,signal to=west] at (0.25,-1.25) {};
  \node (Etext) at (0.1,-1.25) {$E$};
  \node (Ex) [latent_var] at (-2,-1.25) {$E(\x)$};
  \draw[arrow] (z) -> (G);
  \draw[arrow] (G) -> (Gz);
  \draw[arrow] (x) -> (E);
  \draw[arrow] (E) -> (Ex);
  \begin{scope}[on background layer]
    \node[box,fill=blue!10,fit=(z) (Ex), label={above:features}] (container) {};
    \node[box,fill=green!10,fit=(x) (Gz), label={above:data}] (container) {};
  \end{scope}
  \node (Gzz) [group_var] at (4,0.5) {$G(\z), \z$};
  \node (xEx) [group_var] at (4,-0.5) {$\x, E(\x)$};
  \draw[arrow] (z) |- (Gzz);
  \draw[arrow] (Gz) |- (Gzz);
  \draw[arrow] (x) |- (xEx);
  \draw[arrow] (Ex) |- (xEx);
  \node (D) [func] at (6,0) {};
  \node (Dtext) at (6.15,0) {$D$};
  \draw[arrow] (Gzz) to[-|-] ($(D.north west)!0.5!(D.west)$);
  \draw[arrow] (xEx) to[-|-] ($(D.south west)!0.5!(D.west)$);
  
  \node (Py) [latent_var] at (8,0) {$P(y)$};
  \draw[arrow] (D) -- (Py);
 \end{tikzpicture}
}

\caption{The structure of Bidirectional Generative Adversarial Networks (BiGAN).}
\label{fig:concept}
\end{figure}

%% file: table_mnist.tex
\begin{table}
\centering
\begin{tabular}{cccccc}
\toprule
BiGAN & $D$ & LR & JLR & AE ($\ell_2$) & AE ($\ell_1$) \\
\midrule
97.39 & 97.30 & 97.44 & 97.13 & 97.58 & 97.63 \\
\bottomrule
\end{tabular}
\caption{
One Nearest Neighbors (1NN) classification accuracy (\%)
on the permutation-invariant MNIST~\citep{mnist}
test set in the feature space learned by BiGAN, Latent Regressor (LR), Joint Latent Regressor (JLR), and an autoencoder (AE) using an $\ell_1$ or $\ell_2$ distance.}
\lbltbl{mnist}
\end{table}

%% file: qual_mnist.tex
\begin{figure}
\centering
\begingroup
\renewcommand{\arraystretch}{1.7}
\begin{tabular}{rc@{}c@{}c@{}c@{}c@{}c@{}c@{}c@{}c@{}c@{}c@{}c@{}c@{}c@{}c@{}c@{}c@{}c@{}c@{}c@{}}
$G(\z)\phantom{)}$ &
\includegraphics[width=0.037\linewidth,valign=c]{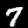} &
\includegraphics[width=0.037\linewidth,valign=c]{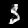} &
\includegraphics[width=0.037\linewidth,valign=c]{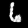} &
\includegraphics[width=0.037\linewidth,valign=c]{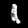} &
\includegraphics[width=0.037\linewidth,valign=c]{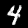} &
\includegraphics[width=0.037\linewidth,valign=c]{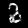} &
\includegraphics[width=0.037\linewidth,valign=c]{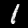} &
\includegraphics[width=0.037\linewidth,valign=c]{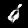} &
\includegraphics[width=0.037\linewidth,valign=c]{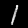} &
\includegraphics[width=0.037\linewidth,valign=c]{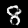} &
\includegraphics[width=0.037\linewidth,valign=c]{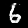} &
\includegraphics[width=0.037\linewidth,valign=c]{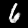} &
\includegraphics[width=0.037\linewidth,valign=c]{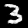} &
\includegraphics[width=0.037\linewidth,valign=c]{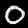} &
\includegraphics[width=0.037\linewidth,valign=c]{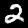} &
\includegraphics[width=0.037\linewidth,valign=c]{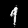} &
\includegraphics[width=0.037\linewidth,valign=c]{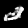} &
\includegraphics[width=0.037\linewidth,valign=c]{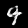} &
\includegraphics[width=0.037\linewidth,valign=c]{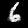} &
\includegraphics[width=0.037\linewidth,valign=c]{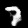} \\
\midrule
$\x\phantom{))}$ &
\includegraphics[width=0.037\linewidth,valign=c]{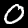} &
\includegraphics[width=0.037\linewidth,valign=c]{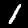} &
\includegraphics[width=0.037\linewidth,valign=c]{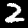} &
\includegraphics[width=0.037\linewidth,valign=c]{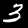} &
\includegraphics[width=0.037\linewidth,valign=c]{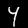} &
\includegraphics[width=0.037\linewidth,valign=c]{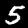} &
\includegraphics[width=0.037\linewidth,valign=c]{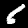} &
\includegraphics[width=0.037\linewidth,valign=c]{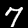} &
\includegraphics[width=0.037\linewidth,valign=c]{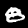} &
\includegraphics[width=0.037\linewidth,valign=c]{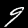} &
\includegraphics[width=0.037\linewidth,valign=c]{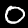} &
\includegraphics[width=0.037\linewidth,valign=c]{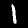} &
\includegraphics[width=0.037\linewidth,valign=c]{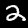} &
\includegraphics[width=0.037\linewidth,valign=c]{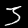} &
\includegraphics[width=0.037\linewidth,valign=c]{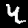} &
\includegraphics[width=0.037\linewidth,valign=c]{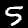} &
\includegraphics[width=0.037\linewidth,valign=c]{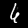} &
\includegraphics[width=0.037\linewidth,valign=c]{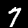} &
\includegraphics[width=0.037\linewidth,valign=c]{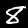} &
\includegraphics[width=0.037\linewidth,valign=c]{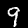} \\
$G(E(\x))$ &
\includegraphics[width=0.037\linewidth,valign=c]{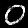} &
\includegraphics[width=0.037\linewidth,valign=c]{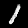} &
\includegraphics[width=0.037\linewidth,valign=c]{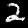} &
\includegraphics[width=0.037\linewidth,valign=c]{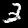} &
\includegraphics[width=0.037\linewidth,valign=c]{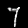} &
\includegraphics[width=0.037\linewidth,valign=c]{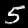} &
\includegraphics[width=0.037\linewidth,valign=c]{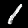} &
\includegraphics[width=0.037\linewidth,valign=c]{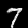} &
\includegraphics[width=0.037\linewidth,valign=c]{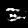} &
\includegraphics[width=0.037\linewidth,valign=c]{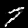} &
\includegraphics[width=0.037\linewidth,valign=c]{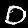} &
\includegraphics[width=0.037\linewidth,valign=c]{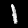} &
\includegraphics[width=0.037\linewidth,valign=c]{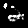} &
\includegraphics[width=0.037\linewidth,valign=c]{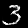} &
\includegraphics[width=0.037\linewidth,valign=c]{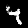} &
\includegraphics[width=0.037\linewidth,valign=c]{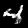} &
\includegraphics[width=0.037\linewidth,valign=c]{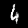} &
\includegraphics[width=0.037\linewidth,valign=c]{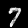} &
\includegraphics[width=0.037\linewidth,valign=c]{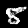} &
\includegraphics[width=0.037\linewidth,valign=c]{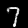} \\
\end{tabular}
\endgroup
\caption{
Qualitative results for permutation-invariant MNIST BiGAN training,
including generator samples $G(\z)$,
real data $\x$,
and corresponding reconstructions $G(E(\x))$.
}
\lblfig{mnistqual}
\end{figure}

%% file: imagenet_filters.tex
\begin{figure*}
\centering

\begin{subfigure}[b]{.15\linewidth}
  \centering
  \includegraphics[width=0.9\linewidth]{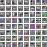}
  \caption*{$D$}
\end{subfigure}
\begin{subfigure}[b]{.4\linewidth}
  \centering
  \includegraphics[width=0.9\linewidth]{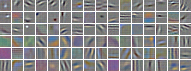}
  \caption*{$E$}
\end{subfigure}
\begin{subfigure}[b]{.4\linewidth}
  \centering
  \includegraphics[width=0.9\linewidth]{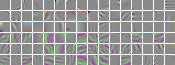}
  \caption*{\citet{jigsaw}}
\end{subfigure}

\begin{subfigure}[b]{.15\columnwidth}
  \centering
  \includegraphics[width=0.9\linewidth]{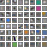}
  \caption*{$G$}
\end{subfigure}
\begin{subfigure}[b]{.4\linewidth}
  \centering
  \includegraphics[width=0.9\linewidth]{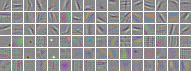}
  \caption*{AlexNet-based $D$}
\end{subfigure}
\begin{subfigure}[b]{.4\linewidth}
  \centering
  \includegraphics[width=0.9\linewidth]{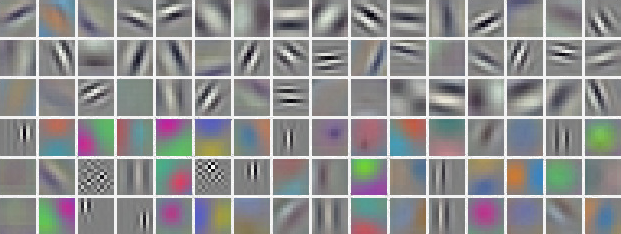}
  \caption*{\citet{supervision}}
\end{subfigure}
\caption{
The convolutional filters learned by the three modules ($D$, $G$, and $E$) of a BiGAN (left, top-middle)
trained on the ImageNet~\citep{imagenet} database.
We compare with the filters learned by a discriminator $D$ trained with the same architecture (bottom-middle),
as well as the filters reported by~\citet{jigsaw},
and by~\citet{supervision} for fully supervised ImageNet training
(right).
}
\label{fig:filters}
\end{figure*}

%% file: qual_imagenet.tex
\begin{figure}
\centering
\begingroup
\renewcommand{\arraystretch}{1.7}
\begin{tabular}{rc@{}c@{}c@{}c@{}c@{}c@{}c@{}c@{}c@{}c@{}c@{}c@{}c@{}c@{}c@{}c@{}c@{}c@{}c@{}c@{}}
\multirow{3}{*}{$G(\z)\phantom{)}$} &
\includegraphics[width=0.037\linewidth,valign=c]{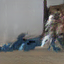} &
\includegraphics[width=0.037\linewidth,valign=c]{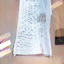} &
\includegraphics[width=0.037\linewidth,valign=c]{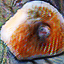} &
\includegraphics[width=0.037\linewidth,valign=c]{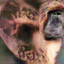} &
\includegraphics[width=0.037\linewidth,valign=c]{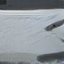} &
\includegraphics[width=0.037\linewidth,valign=c]{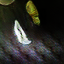} &
\includegraphics[width=0.037\linewidth,valign=c]{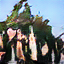} &
\includegraphics[width=0.037\linewidth,valign=c]{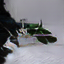} &
\includegraphics[width=0.037\linewidth,valign=c]{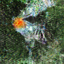} &
\includegraphics[width=0.037\linewidth,valign=c]{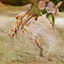} &
\includegraphics[width=0.037\linewidth,valign=c]{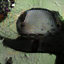} &
\includegraphics[width=0.037\linewidth,valign=c]{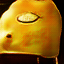} &
\includegraphics[width=0.037\linewidth,valign=c]{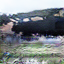} &
\includegraphics[width=0.037\linewidth,valign=c]{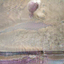} &
\includegraphics[width=0.037\linewidth,valign=c]{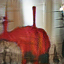} &
\includegraphics[width=0.037\linewidth,valign=c]{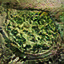} &
\includegraphics[width=0.037\linewidth,valign=c]{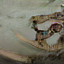} &
\includegraphics[width=0.037\linewidth,valign=c]{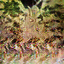} &
\includegraphics[width=0.037\linewidth,valign=c]{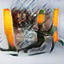} &
\includegraphics[width=0.037\linewidth,valign=c]{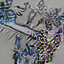} \\
&
\includegraphics[width=0.037\linewidth,valign=c]{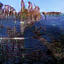} &
\includegraphics[width=0.037\linewidth,valign=c]{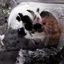} &
\includegraphics[width=0.037\linewidth,valign=c]{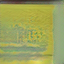} &
\includegraphics[width=0.037\linewidth,valign=c]{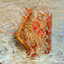} &
\includegraphics[width=0.037\linewidth,valign=c]{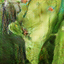} &
\includegraphics[width=0.037\linewidth,valign=c]{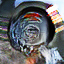} &
\includegraphics[width=0.037\linewidth,valign=c]{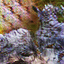} &
\includegraphics[width=0.037\linewidth,valign=c]{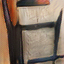} &
\includegraphics[width=0.037\linewidth,valign=c]{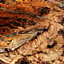} &
\includegraphics[width=0.037\linewidth,valign=c]{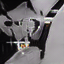} &
\includegraphics[width=0.037\linewidth,valign=c]{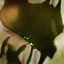} &
\includegraphics[width=0.037\linewidth,valign=c]{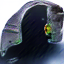} &
\includegraphics[width=0.037\linewidth,valign=c]{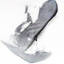} &
\includegraphics[width=0.037\linewidth,valign=c]{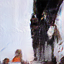} &
\includegraphics[width=0.037\linewidth,valign=c]{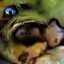} &
\includegraphics[width=0.037\linewidth,valign=c]{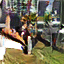} &
\includegraphics[width=0.037\linewidth,valign=c]{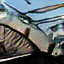} &
\includegraphics[width=0.037\linewidth,valign=c]{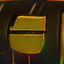} &
\includegraphics[width=0.037\linewidth,valign=c]{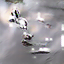} &
\includegraphics[width=0.037\linewidth,valign=c]{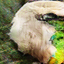} \\
&
\includegraphics[width=0.037\linewidth,valign=c]{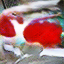} &
\includegraphics[width=0.037\linewidth,valign=c]{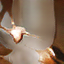} &
\includegraphics[width=0.037\linewidth,valign=c]{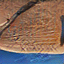} &
\includegraphics[width=0.037\linewidth,valign=c]{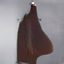} &
\includegraphics[width=0.037\linewidth,valign=c]{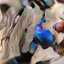} &
\includegraphics[width=0.037\linewidth,valign=c]{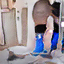} &
\includegraphics[width=0.037\linewidth,valign=c]{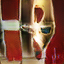} &
\includegraphics[width=0.037\linewidth,valign=c]{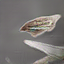} &
\includegraphics[width=0.037\linewidth,valign=c]{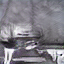} &
\includegraphics[width=0.037\linewidth,valign=c]{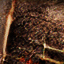} &
\includegraphics[width=0.037\linewidth,valign=c]{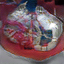} &
\includegraphics[width=0.037\linewidth,valign=c]{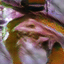} &
\includegraphics[width=0.037\linewidth,valign=c]{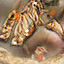} &
\includegraphics[width=0.037\linewidth,valign=c]{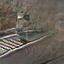} &
\includegraphics[width=0.037\linewidth,valign=c]{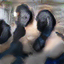} &
\includegraphics[width=0.037\linewidth,valign=c]{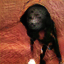} &
\includegraphics[width=0.037\linewidth,valign=c]{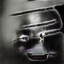} &
\includegraphics[width=0.037\linewidth,valign=c]{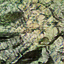} &
\includegraphics[width=0.037\linewidth,valign=c]{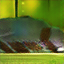} &
\includegraphics[width=0.037\linewidth,valign=c]{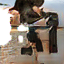} \\
\midrule
$\x\phantom{))}$ &
\includegraphics[width=0.037\linewidth,valign=c]{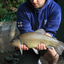} &
\includegraphics[width=0.037\linewidth,valign=c]{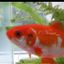} &
\includegraphics[width=0.037\linewidth,valign=c]{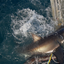} &
\includegraphics[width=0.037\linewidth,valign=c]{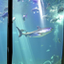} &
\includegraphics[width=0.037\linewidth,valign=c]{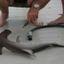} &
\includegraphics[width=0.037\linewidth,valign=c]{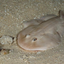} &
\includegraphics[width=0.037\linewidth,valign=c]{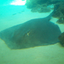} &
\includegraphics[width=0.037\linewidth,valign=c]{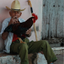} &
\includegraphics[width=0.037\linewidth,valign=c]{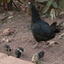} &
\includegraphics[width=0.037\linewidth,valign=c]{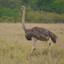} &
\includegraphics[width=0.037\linewidth,valign=c]{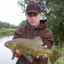} &
\includegraphics[width=0.037\linewidth,valign=c]{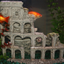} &
\includegraphics[width=0.037\linewidth,valign=c]{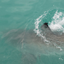} &
\includegraphics[width=0.037\linewidth,valign=c]{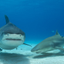} &
\includegraphics[width=0.037\linewidth,valign=c]{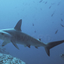} &
\includegraphics[width=0.037\linewidth,valign=c]{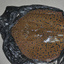} &
\includegraphics[width=0.037\linewidth,valign=c]{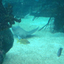} &
\includegraphics[width=0.037\linewidth,valign=c]{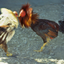} &
\includegraphics[width=0.037\linewidth,valign=c]{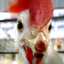} &
\includegraphics[width=0.037\linewidth,valign=c]{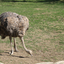} \\
$G(E(\x))$ &
\includegraphics[width=0.037\linewidth,valign=c]{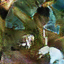} &
\includegraphics[width=0.037\linewidth,valign=c]{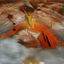} &
\includegraphics[width=0.037\linewidth,valign=c]{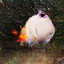} &
\includegraphics[width=0.037\linewidth,valign=c]{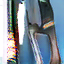} &
\includegraphics[width=0.037\linewidth,valign=c]{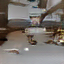} &
\includegraphics[width=0.037\linewidth,valign=c]{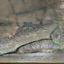} &
\includegraphics[width=0.037\linewidth,valign=c]{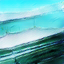} &
\includegraphics[width=0.037\linewidth,valign=c]{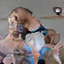} &
\includegraphics[width=0.037\linewidth,valign=c]{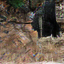} &
\includegraphics[width=0.037\linewidth,valign=c]{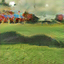} &
\includegraphics[width=0.037\linewidth,valign=c]{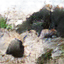} &
\includegraphics[width=0.037\linewidth,valign=c]{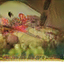} &
\includegraphics[width=0.037\linewidth,valign=c]{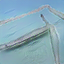} &
\includegraphics[width=0.037\linewidth,valign=c]{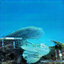} &
\includegraphics[width=0.037\linewidth,valign=c]{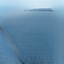} &
\includegraphics[width=0.037\linewidth,valign=c]{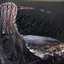} &
\includegraphics[width=0.037\linewidth,valign=c]{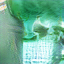} &
\includegraphics[width=0.037\linewidth,valign=c]{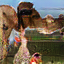} &
\includegraphics[width=0.037\linewidth,valign=c]{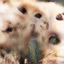} &
\includegraphics[width=0.037\linewidth,valign=c]{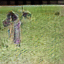} \\
\midrule
$\x\phantom{))}$ &
\includegraphics[width=0.037\linewidth,valign=c]{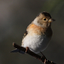} &
\includegraphics[width=0.037\linewidth,valign=c]{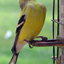} &
\includegraphics[width=0.037\linewidth,valign=c]{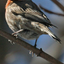} &
\includegraphics[width=0.037\linewidth,valign=c]{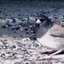} &
\includegraphics[width=0.037\linewidth,valign=c]{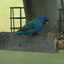} &
\includegraphics[width=0.037\linewidth,valign=c]{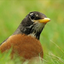} &
\includegraphics[width=0.037\linewidth,valign=c]{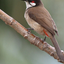} &
\includegraphics[width=0.037\linewidth,valign=c]{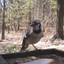} &
\includegraphics[width=0.037\linewidth,valign=c]{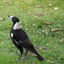} &
\includegraphics[width=0.037\linewidth,valign=c]{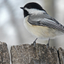} &
\includegraphics[width=0.037\linewidth,valign=c]{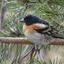} &
\includegraphics[width=0.037\linewidth,valign=c]{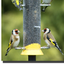} &
\includegraphics[width=0.037\linewidth,valign=c]{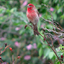} &
\includegraphics[width=0.037\linewidth,valign=c]{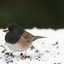} &
\includegraphics[width=0.037\linewidth,valign=c]{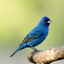} &
\includegraphics[width=0.037\linewidth,valign=c]{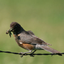} &
\includegraphics[width=0.037\linewidth,valign=c]{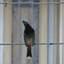} &
\includegraphics[width=0.037\linewidth,valign=c]{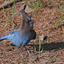} &
\includegraphics[width=0.037\linewidth,valign=c]{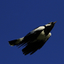} &
\includegraphics[width=0.037\linewidth,valign=c]{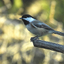} \\
$G(E(\x))$ &
\includegraphics[width=0.037\linewidth,valign=c]{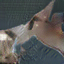} &
\includegraphics[width=0.037\linewidth,valign=c]{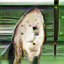} &
\includegraphics[width=0.037\linewidth,valign=c]{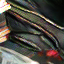} &
\includegraphics[width=0.037\linewidth,valign=c]{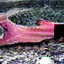} &
\includegraphics[width=0.037\linewidth,valign=c]{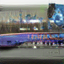} &
\includegraphics[width=0.037\linewidth,valign=c]{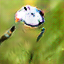} &
\includegraphics[width=0.037\linewidth,valign=c]{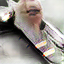} &
\includegraphics[width=0.037\linewidth,valign=c]{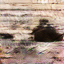} &
\includegraphics[width=0.037\linewidth,valign=c]{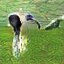} &
\includegraphics[width=0.037\linewidth,valign=c]{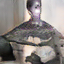} &
\includegraphics[width=0.037\linewidth,valign=c]{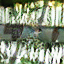} &
\includegraphics[width=0.037\linewidth,valign=c]{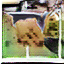} &
\includegraphics[width=0.037\linewidth,valign=c]{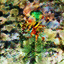} &
\includegraphics[width=0.037\linewidth,valign=c]{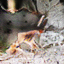} &
\includegraphics[width=0.037\linewidth,valign=c]{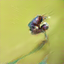} &
\includegraphics[width=0.037\linewidth,valign=c]{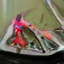} &
\includegraphics[width=0.037\linewidth,valign=c]{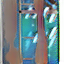} &
\includegraphics[width=0.037\linewidth,valign=c]{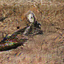} &
\includegraphics[width=0.037\linewidth,valign=c]{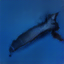} &
\includegraphics[width=0.037\linewidth,valign=c]{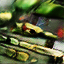} \\
\midrule
$\x\phantom{))}$ &
\includegraphics[width=0.037\linewidth,valign=c]{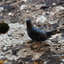} &
\includegraphics[width=0.037\linewidth,valign=c]{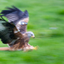} &
\includegraphics[width=0.037\linewidth,valign=c]{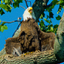} &
\includegraphics[width=0.037\linewidth,valign=c]{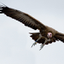} &
\includegraphics[width=0.037\linewidth,valign=c]{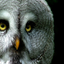} &
\includegraphics[width=0.037\linewidth,valign=c]{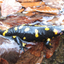} &
\includegraphics[width=0.037\linewidth,valign=c]{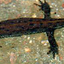} &
\includegraphics[width=0.037\linewidth,valign=c]{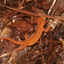} &
\includegraphics[width=0.037\linewidth,valign=c]{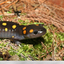} &
\includegraphics[width=0.037\linewidth,valign=c]{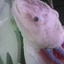} &
\includegraphics[width=0.037\linewidth,valign=c]{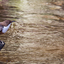} &
\includegraphics[width=0.037\linewidth,valign=c]{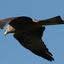} &
\includegraphics[width=0.037\linewidth,valign=c]{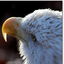} &
\includegraphics[width=0.037\linewidth,valign=c]{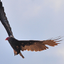} &
\includegraphics[width=0.037\linewidth,valign=c]{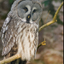} &
\includegraphics[width=0.037\linewidth,valign=c]{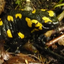} &
\includegraphics[width=0.037\linewidth,valign=c]{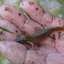} &
\includegraphics[width=0.037\linewidth,valign=c]{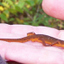} &
\includegraphics[width=0.037\linewidth,valign=c]{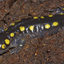} &
\includegraphics[width=0.037\linewidth,valign=c]{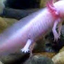} \\
$G(E(\x))$ &
\includegraphics[width=0.037\linewidth,valign=c]{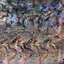} &
\includegraphics[width=0.037\linewidth,valign=c]{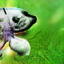} &
\includegraphics[width=0.037\linewidth,valign=c]{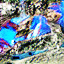} &
\includegraphics[width=0.037\linewidth,valign=c]{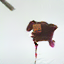} &
\includegraphics[width=0.037\linewidth,valign=c]{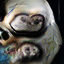} &
\includegraphics[width=0.037\linewidth,valign=c]{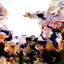} &
\includegraphics[width=0.037\linewidth,valign=c]{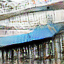} &
\includegraphics[width=0.037\linewidth,valign=c]{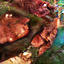} &
\includegraphics[width=0.037\linewidth,valign=c]{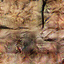} &
\includegraphics[width=0.037\linewidth,valign=c]{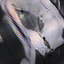} &
\includegraphics[width=0.037\linewidth,valign=c]{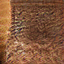} &
\includegraphics[width=0.037\linewidth,valign=c]{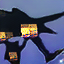} &
\includegraphics[width=0.037\linewidth,valign=c]{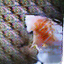} &
\includegraphics[width=0.037\linewidth,valign=c]{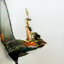} &
\includegraphics[width=0.037\linewidth,valign=c]{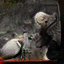} &
\includegraphics[width=0.037\linewidth,valign=c]{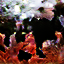} &
\includegraphics[width=0.037\linewidth,valign=c]{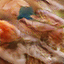} &
\includegraphics[width=0.037\linewidth,valign=c]{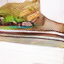} &
\includegraphics[width=0.037\linewidth,valign=c]{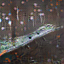} &
\includegraphics[width=0.037\linewidth,valign=c]{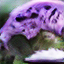} \\
\end{tabular}
\endgroup
\caption{
Qualitative results for ImageNet BiGAN training,
including generator samples $G(\z)$,
real data $\x$,
and corresponding reconstructions $G(E(\x))$.
}
\lblfig{imagenetqual}
\end{figure}

%% file: table_imagenet_class.tex
\begin{table}
\centering
\begin{tabular}{lccccc}
\toprule
& conv1 & conv2 & conv3 & conv4 & conv5 \\
\midrule
Random \citep{jigsaw}            & 48.5 & 41.0 & 34.8 & 27.1 & 12.0 \\
\citet{xiaolong}                 & 51.8 & 46.9 & 42.8 & 38.8 & 29.8 \\
\citet{carl}                     & 53.1 & 47.6 & 48.7 & \bf{45.6} & 30.4 \\
\gray{\citet{jigsaw}}*           & \gray{57.1} & \gray{56.0} & \gray{52.4} & \gray{48.3} & \gray{38.1} \\
\midrule
BiGAN (ours)                     & \bf{56.2} & \bf{54.4} & \bf{49.4} & 43.9 & 33.3 \\
BiGAN, $112\times112$ $E$ (ours) & 55.3 & 53.2 & 49.3 & 44.4 & \bf{34.8} \\
\bottomrule
\end{tabular}
\caption{
Classification accuracy (\%) for the ImageNet LSVRC~\citep{imagenet} validation set with various portions of the network frozen, or reinitialized and trained from scratch, following the evaluation from~\citet{jigsaw}.
In, e.g., the \textit{conv3} column, the first three layers -- conv1 through conv3 --
are transferred and frozen, and the last layers --
conv4, conv5, and fully connected layers --
are reinitialized and trained fully supervised for ImageNet classification.
BiGAN is competitive with these contemporary visual feature learning methods, despite its generality.
(*Results from \citet{jigsaw} are not directly comparable to those of the other methods
as a different base convnet architecture with larger intermediate feature maps is used.)
}
\lbltbl{imagenetclass}
\end{table}

%% file: table_voc_all.tex
\begin{table}
\small
\centering
\begin{tabular}{rlccccc}
\toprule
& & \multicolumn{3}{c}{} & \textit{FRCN} & \textit{FCN} \\
& & \multicolumn{3}{c}{Classification} & Detection & Segmentation \\
& & \multicolumn{3}{c}{(\% mAP)} & (\% mAP) & (\% mIU) \\

& trained layers & fc8 & fc6-8 & all & all & all \\
\midrule
\multirow{1}{*}{ sup. }
& ImageNet \citep{supervision}       & \bf{77.0} & \bf{78.8} & \bf{78.3} & \bf{56.8} & \bf{48.0} \\
\midrule
\multirow{4}{*}{ self-sup. }
& \citet{pulkit}                     & 31.2 & 31.0 & 54.2 & 43.9 &  -   \\
& \citet{deepak}                     & 30.5 & 34.6 & 56.5 & 44.5 & \bf{30.0} \\
& \citet{xiaolong}                   & 28.4 & \bf{55.6} & 63.1 & 47.4 &  -   \\
& \citet{carl}                       & \bf{44.7} & 55.1 & \bf{65.3} & \bf{51.1} &  -   \\
\midrule
\multirow{7}{*}{ unsup. }
& $k$-means \citep{datadep}          & 32.0 & 39.2 & 56.6 & 45.6 & 32.6 \\
& Discriminator ($D$)                & 30.7 & 40.5 & 56.4 &  -   & - \\
& Latent Regressor (LR)              & 36.9 & 47.9 & 57.1 &  -   & - \\
& Joint LR                           & 37.1 & 47.9 & 56.5 &  -   & - \\
& Autoencoder ($\ell_2$)             & 24.8 & 16.0 & 53.8 & 41.9 & - \\
& BiGAN (ours)                       & 37.5 & 48.7 & 58.9 & 46.2 & 34.9 \\
& BiGAN, $112\times112$ $E$ (ours)  & \bf{41.7} & \bf{52.5} & \bf{60.3} & \bf{46.9} & \bf{35.2} \\
\bottomrule
\end{tabular}
\caption{
Classification and Fast R-CNN~\citep{fastrcnn} detection results for the PASCAL VOC 2007~\citep{pascal} test set, and FCN~\citep{fcn} segmentation results on the PASCAL VOC 2012 validation set,
under the standard mean average precision (mAP) or mean intersection over union (mIU) metrics for each task.
Classification models are trained with various portions of the \textit{AlexNet}~\citep{supervision} model frozen.
In the \textit{fc8} column, only the linear classifier (a multinomial logistic regression) is learned -- in the case of BiGAN, on top of randomly initialized fully connected (FC) layers \textit{fc6} and \textit{fc7}.
In the \textit{fc6-8} column, all three FC layers are trained fully supervised with all convolution layers frozen.
Finally, in the \textit{all} column, the entire network is ``fine-tuned''.
BiGAN outperforms other unsupervised (\textit{unsup.}) feature learning approaches, including the GAN-based baselines described in \refsec{baselines},
and despite its generality, is competitive with contemporary self-supervised (\textit{self-sup.}) feature learning approaches specific to the visual domain.
}
\lbltbl{vocclass}
\lbltbl{vocloc}
\end{table}

%% file: nn_imagenet.tex
\begin{figure}
\centering
\begin{tabular}{c|cccc}
\hline
Query & \#1 & \#2 & \#3 & \#4 \\
\hline
\\
\includegraphics[width=0.16\linewidth,valign=c]{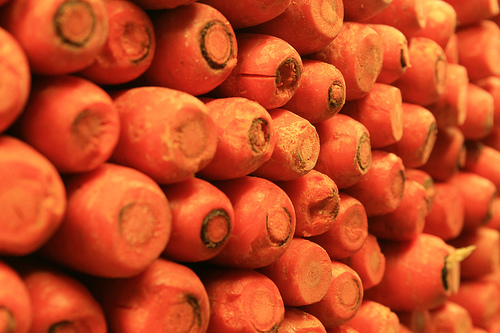} &
\includegraphics[width=0.16\linewidth,valign=c]{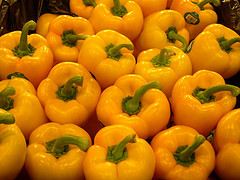} &
\includegraphics[width=0.16\linewidth,valign=c]{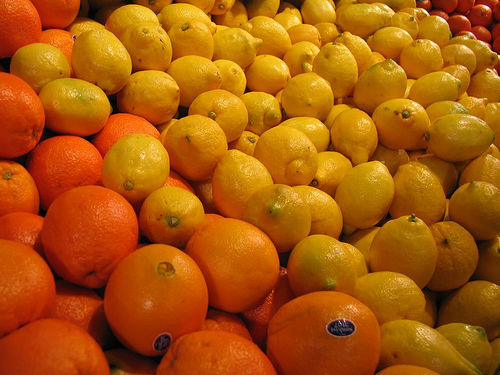} &
\includegraphics[width=0.16\linewidth,valign=c]{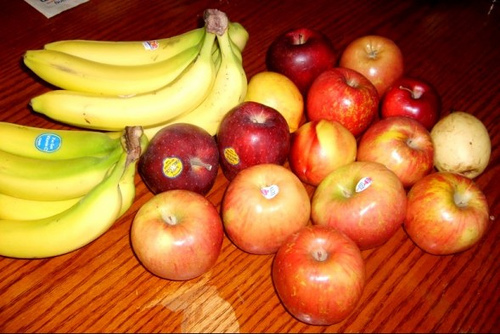} &
\includegraphics[width=0.16\linewidth,valign=c]{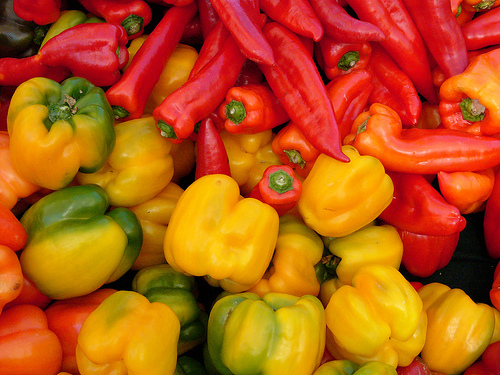} \\
\\
\includegraphics[width=0.16\linewidth,valign=c]{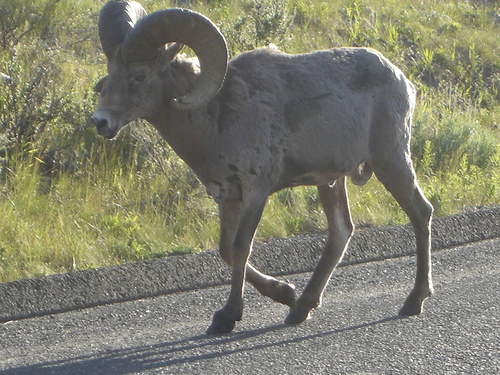} &
\includegraphics[width=0.16\linewidth,valign=c]{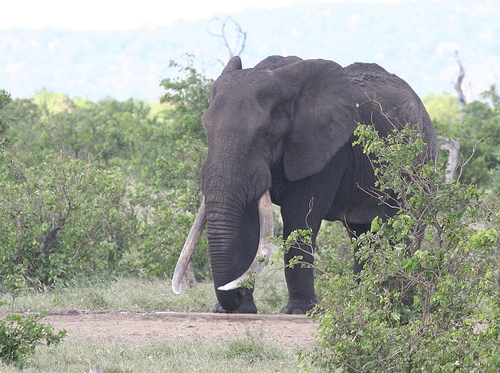} &
\includegraphics[width=0.16\linewidth,valign=c]{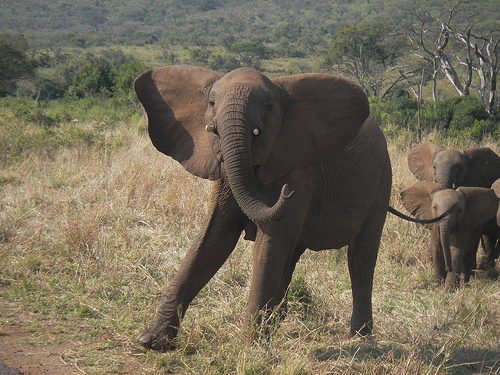} &
\includegraphics[width=0.16\linewidth,valign=c]{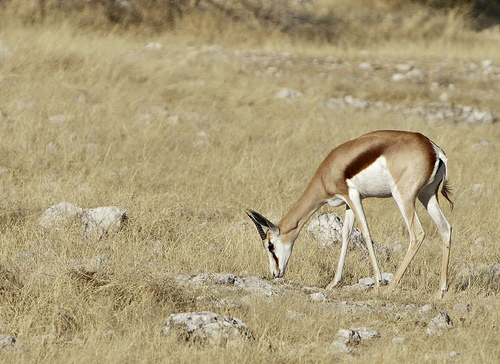} &
\includegraphics[width=0.16\linewidth,valign=c]{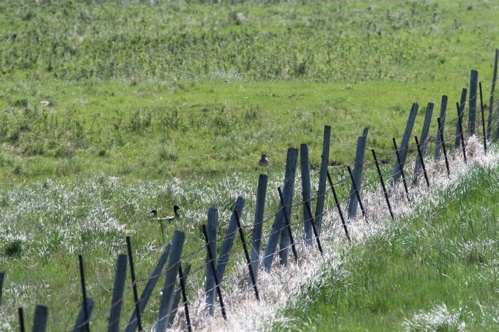} \\
\\
\includegraphics[width=0.16\linewidth,valign=c]{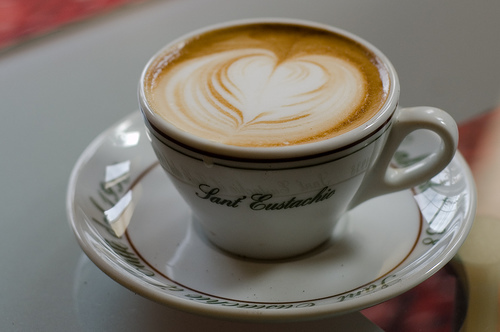} &
\includegraphics[width=0.16\linewidth,valign=c]{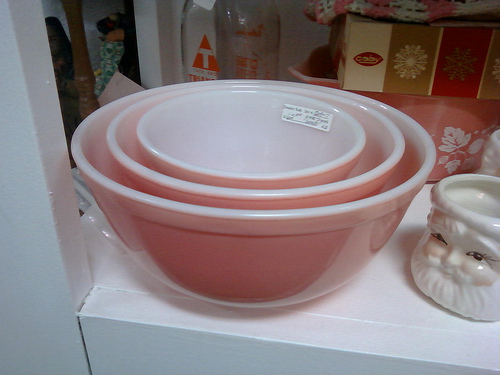} &
\includegraphics[width=0.16\linewidth,valign=c]{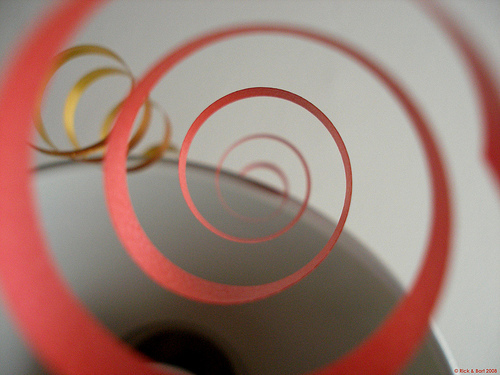} &
\includegraphics[width=0.16\linewidth,valign=c]{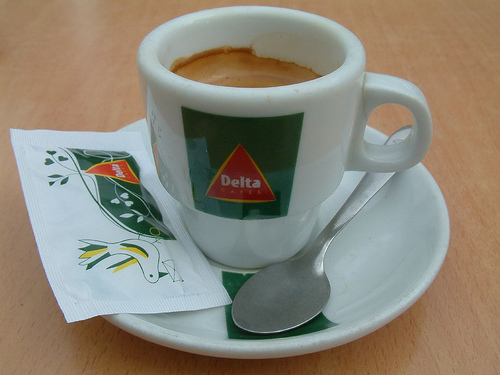} &
\includegraphics[width=0.16\linewidth,valign=c]{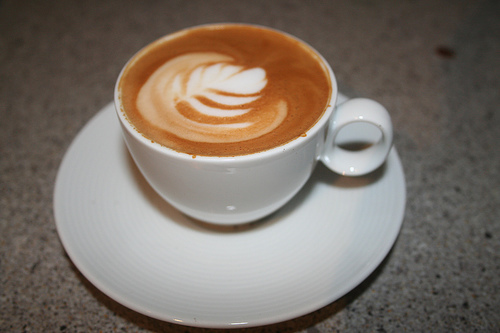} \\
\\
\includegraphics[width=0.16\linewidth,valign=c]{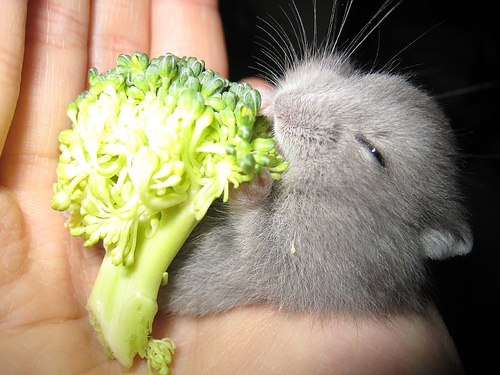} &
\includegraphics[width=0.16\linewidth,valign=c]{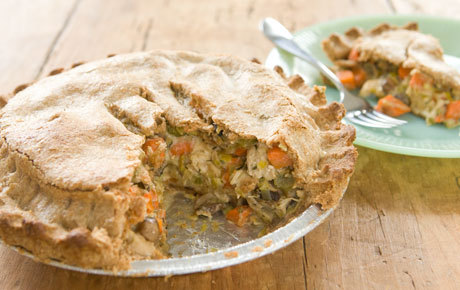} &
\includegraphics[width=0.16\linewidth,valign=c]{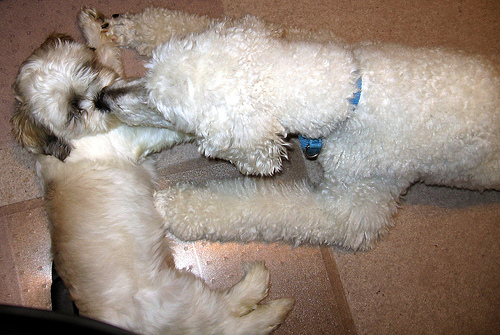} &
\includegraphics[width=0.16\linewidth,valign=c]{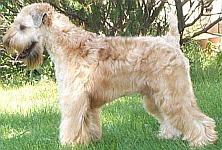} &
\includegraphics[width=0.16\linewidth,valign=c]{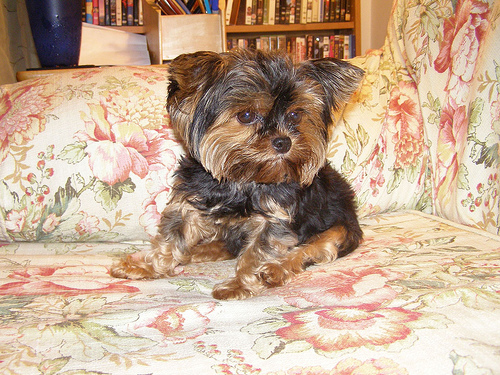} \\
\\
\includegraphics[width=0.16\linewidth,valign=c]{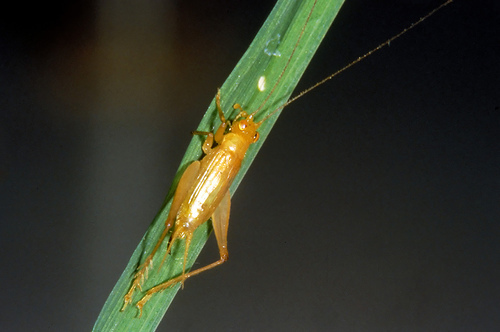} &
\includegraphics[width=0.16\linewidth,valign=c]{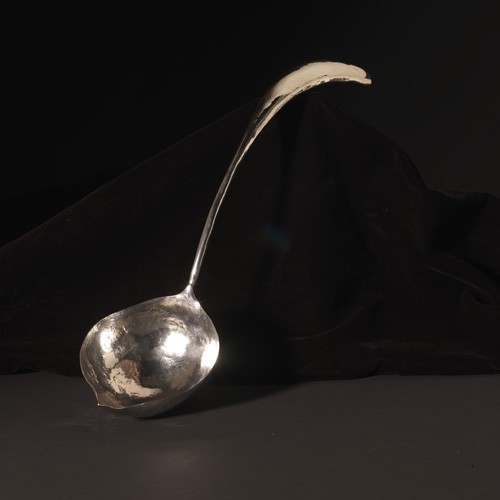} &
\includegraphics[width=0.16\linewidth,valign=c]{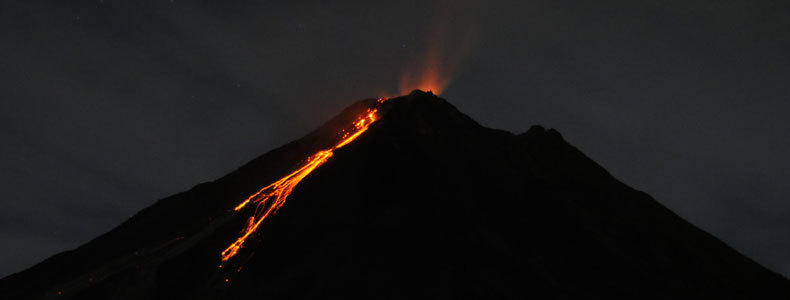} &
\includegraphics[width=0.16\linewidth,valign=c]{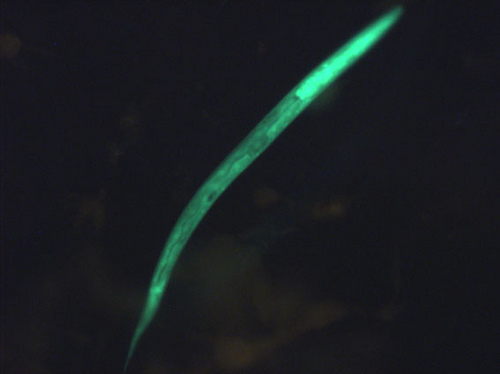} &
\includegraphics[width=0.16\linewidth,valign=c]{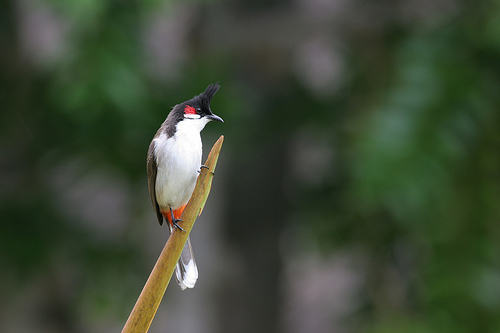} \\
\\
\includegraphics[width=0.16\linewidth,valign=c]{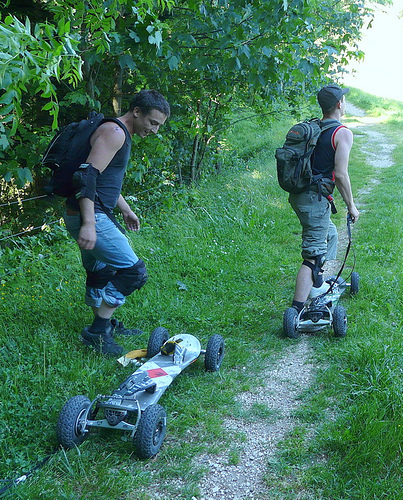} &
\includegraphics[width=0.16\linewidth,valign=c]{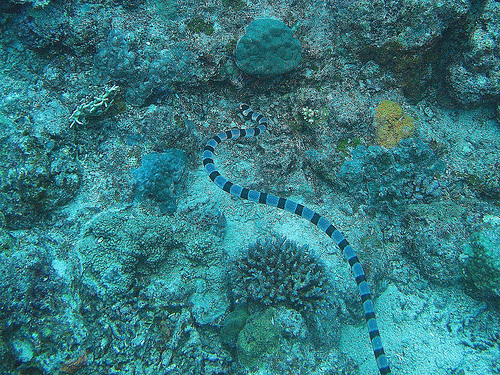} &
\includegraphics[width=0.16\linewidth,valign=c]{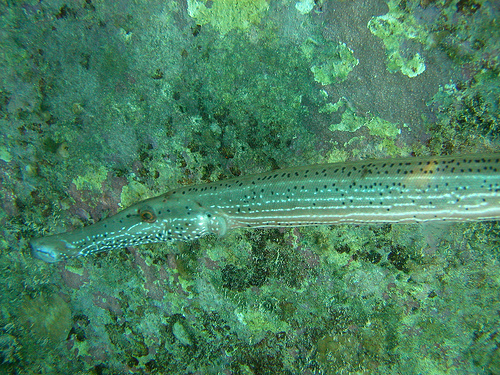} &
\includegraphics[width=0.16\linewidth,valign=c]{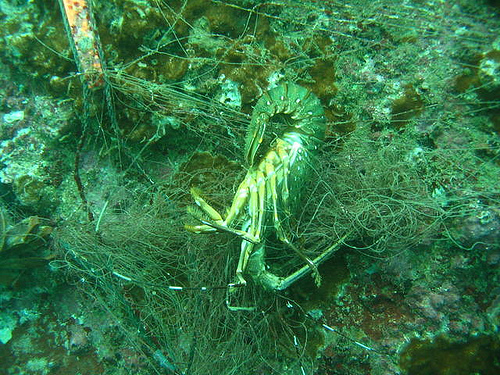} &
\includegraphics[width=0.16\linewidth,valign=c]{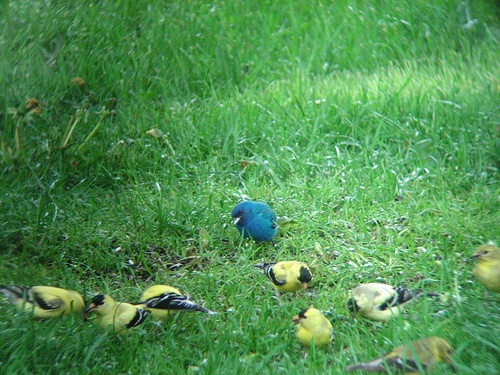} \\
\hline
\end{tabular}
\caption{
For the query images used in~\citet{datadep} (left),
nearest neighbors (by minimum cosine distance) from the ImageNet LSVRC~\citep{imagenet} training set in the \textit{fc6} feature space of the ImageNet-trained BiGAN encoder $E$.
(The \textit{fc6} weights are set randomly; this space is a random projection of the learned \textit{conv5} feature space.)
}
\lblfig{imagenetneighbors}
\end{figure}